\documentclass[twoside]{article} 
\usepackage[accepted]{aistats2018}

\usepackage{fancyhdr}
\usepackage{multibib}
\usepackage{times}
\usepackage{subfigure}
\usepackage{amsmath}
\usepackage{amsthm}
\usepackage{amsfonts}
\usepackage{amssymb}
\usepackage{graphicx}
\usepackage{hyperref}
\usepackage{natbib}
\usepackage{algorithmic}
\usepackage{algorithm}

\usepackage{color}
\usepackage{xcolor}
\usepackage{hyperref}
\usepackage{graphicx}
\usepackage{multirow}
\newcommand{\dkls}[3]{\mathbb{D}_{KL}^{#1}[#2 \, \|\, #3]}

\newcommand{\sg}[2]{\widehat{\nabla}_{#2} #1}

\newcommand\cut[1]{}

\usepackage{algorithm}

\newcommand{\squishlist}{
   \begin{list}{$\bullet$}
    { \setlength{\itemsep}{0pt}      \setlength{\parsep}{3pt}
      \setlength{\topsep}{3pt}       \setlength{\partopsep}{0pt}
      \setlength{\leftmargin}{1.5em} \setlength{\labelwidth}{1em}
      \setlength{\labelsep}{0.5em} } }

\newcommand{\squishlisttwo}{
   \begin{list}{$\bullet$}
    { \setlength{\itemsep}{0pt}    \setlength{\parsep}{0pt}
      \setlength{\topsep}{0pt}     \setlength{\partopsep}{0pt}
      \setlength{\leftmargin}{2em} \setlength{\labelwidth}{1.5em}
      \setlength{\labelsep}{0.5em} } }

\newcommand{\squishend}{
    \end{list}  }
 
{}
{}
{}
{}
{}

\newcommand{\half}{\mbox{$\frac{1}{2}$}}

\newcommand{\real}{\mbox{$\mathbb{R}$}}

\newcommand{\rnd}[1]{\left(#1\right)}
\newcommand{\sqr}[1]{\left[#1\right]}

\newcommand{\myexpect}{\mathbb{E}}

\newcommand{\gauss}{\mbox{${\cal N}$}}

\newcommand{\prob}{\mbox{$p$}}

\newcommand{\myvec}[1]{\mbox{$\mathbf{#1}$}}
\newcommand{\myvecsym}[1]{\mbox{$\boldsymbol{#1}$}}

\newcommand{\vepsilon}{\mbox{$\myvecsym{\epsilon}$}}
\newcommand{\veta}{\mbox{$\myvecsym{\eta}$}}

\newcommand{\vmu}{\mbox{$\myvecsym{\mu}$}}
\newcommand{\vlambda}{\mbox{$\myvecsym{\lambda}$}}

\newcommand{\vphi}{\mbox{$\myvecsym{\phi}$}}

\newcommand{\vpi}{\mbox{$\myvecsym{\pi}$}}

\newcommand{\vtheta}{\mbox{$\myvecsym{\theta}$}}

\newcommand{\vsigma}{\mbox{$\myvecsym{\sigma}$}}
\newcommand{\vSigma}{\mbox{$\myvecsym{\Sigma}$}}

\newcommand{\va}{\mbox{$\myvec{a}$}}

\newcommand{\vg}{\mbox{$\myvec{g}$}}
\newcommand{\vh}{\mbox{$\myvec{h}$}}

\newcommand{\vm}{\mbox{$\myvec{m}$}}

\newcommand{\vs}{\mbox{$\myvec{s}$}}

\newcommand{\vx}{\mbox{$\myvec{x}$}}

\newcommand{\vy}{\mbox{$\myvec{y}$}}
\newcommand{\vyt}{\mbox{$\myvec{\tilde{y}}$}}
\newcommand{\vz}{\mbox{$\myvec{z}$}}

\newcommand{\vM}{\mbox{$\myvec{M}$}}

\newcommand{\vP}{\mbox{$\myvec{P}$}}

\newcommand{\vY}{\mbox{$\myvec{Y}$}}

\newcommand{\diag}{\mbox{$\mbox{diag}$}}

\newcommand{\ee}{\end{equation}}
\newcommand{\bea}{\begin{eqnarray}}
\newcommand{\eea}{\end{eqnarray}}
\newcommand{\beaa}{\begin{eqnarray*}}
\newcommand{\eeaa}{\end{eqnarray*}}

\newcommand{\argmin}{\mathop{\mathrm{argmin}}}
\newcommand{\argmax}{\mathop{\mathrm{argmax}}}

\newcommand{\bx}{\boldsymbol{x}}
\newcommand{\by}{\boldsymbol{y}}
\newcommand{\bz}{\boldsymbol{z}}

\newcommand{\bs}{\boldsymbol{s}}
\newcommand{\ba}{\boldsymbol{a}}

\newcommand{\bv}{\boldsymbol{v}}

\newcommand{\bmu}{\boldsymbol{\mu}}

\newcommand{\btheta}{\boldsymbol{\theta}}

\newcommand{\bSigma}{\boldsymbol{\Sigma}}

\newcommand{\ds}{\mathrm{d}\boldsymbol{s}}
\newcommand{\da}{\mathrm{d}\boldsymbol{a}}

\usepackage[draft]{todonotes}   % notes showed
\usepackage{mdframed}
\newmdenv[
  topline=false,
  bottomline=false,
  rightline=false,
  skipabove=\topsep,
  ]{siderules}

\begin{document}

\runningtitle{Variational Adaptive-Newton Method}

\runningauthor{Khan, Lin, Tangkaratt, Liu, and Nielsen}

\twocolumn[

\aistatstitle{Variational Adaptive-Newton Method for Explorative Learning}

\aistatsauthor{Mohammad Emtiyaz Khan $\quad\quad$ Wu Lin $\quad\quad$ Voot Tangkaratt $\quad\quad$ Zuozhu Liu $\quad\quad$ Didrik Nielsen}

\aistatsaddress{ Center for Advanced Intelligence Project (AIP), RIKEN, Tokyo, Japan } ]

\begin{abstract}
  We present the Variational Adaptive Newton (VAN) method which is a black-box optimization method especially suitable for explorative-learning tasks such as active learning and reinforcement learning. Similar to Bayesian methods, VAN estimates a distribution that can be used for exploration, but requires  computations that are similar to continuous optimization methods. Our theoretical contribution reveals that VAN is a second-order method that unifies existing methods in distinct fields of continuous optimization, variational inference, and  evolution strategies. Our experimental results show that VAN performs well on a wide-variety of learning tasks. This work presents a general-purpose explorative-learning method that has the potential to improve learning in areas such as active learning and reinforcement learning. 
%Our work in this paper presents a general-purpose learning tool that could bring explorative could be useful for difficult learning problems such as deep reinforcement learning and unsupervised learning.

%Bayesian exploration methods are popular in areas such as active learning and reinforcement learning, but they do not scale well to large-scale problems. How can we design alternative explorative-learning methods that are computationally cheap? In this paper, we propose such a method by using a mirror-descent algorithm that minimizes the variational optimization objective. 
%Just like Bayesian methods, our method computes a distribution that can be used for explorative learning, but it uses computations similar to continuous optimization methods. Our theoretical contribution reveals that our method is a second-order method that unifies existing methods in the distinct fields for continuous optimization, variational inference, and evolution strategies. Through experimental comparisons, we show that our method performs comparably with existing methods for supervised and unsupervised learning, and shows promising results for active learning and reinforcement learning.
%

\end{abstract}

\section{Introduction}
\label{sec:intro}
Throughout our life, we continue to learn about the world by sequentially exploring it. We acquire new experiences using old ones, and use the new ones to learn even more.
How can we design methods that can perform such ``explorative" learning to obtain good generalizations? This is an open question in artificial intelligence and machine learning.

%Humans and other living beings acquire useful skills by sequentially exploring the world and collecting relevant experiences about it. During their lifetimes, we contine to explore the world by using our past experiences to acquire new ones and use them to improve our skills.  
%How can we design algorithms that can mimic such type of learning via exploration? This is an open question in machine learning and artificial intelligence.

%One popular approach is based on Bayesian methods, which have been used to model human cognitive development  \citep{perfors2011tutorial}, and also been applied to a wide variety of practical explorative-learning tasks, e.g., in active learning and Bayesian optimization to select informative data examples \citep{2011arXiv1112.5745H, 2017arXiv170302910G, brochu2010tutorial, fishel2012bayesian}, and in reinforcement learning to learn through interactions \citep{wyatt1998exploration, Strens00abayesian}.
%However, inference in Bayesian methods involves estimation of the posterior distribution which is computationally demanding.

One such approach is based on Bayesian methods. This approach has not only been used as a theoretical model of human cognitive development \citep{perfors2011tutorial} but also been applied to a wide variety of practical explorative-learning tasks, e.g., in active learning and Bayesian optimization to select informative data examples \citep{2011arXiv1112.5745H, 2017arXiv170302910G, brochu2010tutorial, fishel2012bayesian}, and in reinforcement learning to learn through interactions \citep{wyatt1998exploration, Strens00abayesian}.
Unfortunately, Bayesian methods are computationally demanding because computation of the posterior distribution is a difficult task, especially for large-scale problems.
%However, Bayesian methods involve estimation of posterior distributions which is a computationally challenging task.
In contrast, non-Bayesian methods, such as those based on continuous optimization methods, are generally computationally much cheaper, but they cannot directly exploit the mechanisms of Bayesian exploration because they do not estimate the posterior distribution.
This raises the following question: how can we design explorative-learning methods that compute a distribution just like Bayesian methods, but cost similar to optimization methods?

%One approach for such explorative learning is based on the Bayesian posterior distributions of the model parameters. This distribution summarizes the past experiences and can be used for explorations that mimic human learning \citep{perfors2011tutorial}.
%Bayesian methods have been successfully applied to a wide variety of explorative-learning tasks. For example, they have been used in active learning and Bayesian optimization to select data example \citep{2011arXiv1112.5745H, 2017arXiv170302910G, brochu2010tutorial, fishel2012bayesian}. They have also been applied to reinforcement learning and sequential decision-making \citep{wyatt1998exploration, Strens00abayesian}.
%However, computing the posterior distribution is computationally demanding which makes it difficult to apply Bayesian methods to large-scale problems.
%
%On the other hand, non-Bayesian methods based on optimization methods, such as stochastic-gradient descent (SGD), are computationally much cheaper. This is because they only compute a point estimate over model parameters rather than a distribution. 
%However, due to the unavailability of a distribution over parameters, non-Bayesian methods cannot directly exploit the Bayesian exploration mechanisms.
%This raises the following question: how can we design explorative-learning methods that compute a distribution just like Bayesian methods, but cost similar to optimization methods? 

In this paper, we propose such a method.
%In this paper, we propose a method whose computation complexity can be very similar to SGD, but it can also use a distribution to perform explorative learning.
Our method can be used to solve generic unconstrained function-minimization problems\footnote{This is a black-box optimization problem in the sense that we may not have access to an analytical form of the function or its derivatives, but we might be able to approximate them at a point.} that take the following form:
%We consider learning tasks that can be formulated as a function minimization problem,
\begin{align}
   \vtheta^* = \argmin_{\btheta} f(\vtheta), \quad \textrm{ where } \vtheta \in \real^D.
  	\label{eq:minf}
\end{align}
A wide variety of problems in supervised, unsupervised, and reinforcement learning can be formulated in this way.
Instead of directly solving the above problem, our method solves it indirectly by first taking the expectation of $f(\vtheta)$ with respect to an unknown probability distribution $q(\vtheta|\veta)$, and then solving the following minimization problem: 
%Our method solve this problem by using a probability distribution within the Variational Optimization (VO) framework \citep{2012arXiv1212.4507S} where, instead of directly minimizing $f(\vtheta)$, we minimize its expectation show below,
\begin{align}
   \min_{\boldsymbol{\eta}}  \,\, \myexpect_{q(\boldsymbol{\theta}|\boldsymbol{\eta})} \sqr{f(\vtheta)} :=  \mathcal{L}(\veta) .
   \label{eq:VO}
\end{align}
where minimization is done with respect to the parameter $\veta$ of the distribution $q$.
This approach is referred to as the Variational Optimization (VO) approach by \citet{2012arXiv1212.4507S} and can lead us to the minimum $\vtheta^*$ because $\mathcal{L}(\veta)$ is an upper bound on the minimum value of $f$, i.e., $ \min_{\btheta} f(\vtheta) \le \myexpect_{q(\boldsymbol{\theta}|\boldsymbol{\eta})} \sqr{f(\vtheta)}$.
Therefore minimizing $\mathcal{L}(\veta)$ minimizes $f(\vtheta)$, and when the distribution $q$ puts all its mass on $\vtheta^*$, we recover the minimum value. This type of function minimization is commonly used in many areas of stochastic search such as evolution strategies \citep{hansen2001completely, wierstra2008natural}. In our problem context, this formulation is advantageous because it enables learning via exploration, where exploration is facilitated through the distribution $q(\vtheta|\veta)$.

%This approach is referred to as the Variational Optimization (VO) approach by \citet{2012arXiv1212.4507S}, and it can be used to minimize $f$ because the following upper bound property of the expectation: $ \min_{\btheta} f(\vtheta) \le \myexpect_{q(\boldsymbol{\theta}|\boldsymbol{\eta})} \sqr{f(\vtheta)}$. That is, minimizing the expectation of the function also minimizes the function itself, and it can lead us to the minimum of $\vtheta^*$ if the distribution $q$ puts all its mass on
%$\vtheta^*$ upon convergence. This type of reformulation is common in many areas of stochastic search such as evolution strategies \citep{hansen2001completely, wierstra2008natural}. In our problem context, this reformulation is advantageous because it enables explorative learning through the distribution $q(\vtheta|\veta)$.

\begin{figure}[!t]
\includegraphics[]{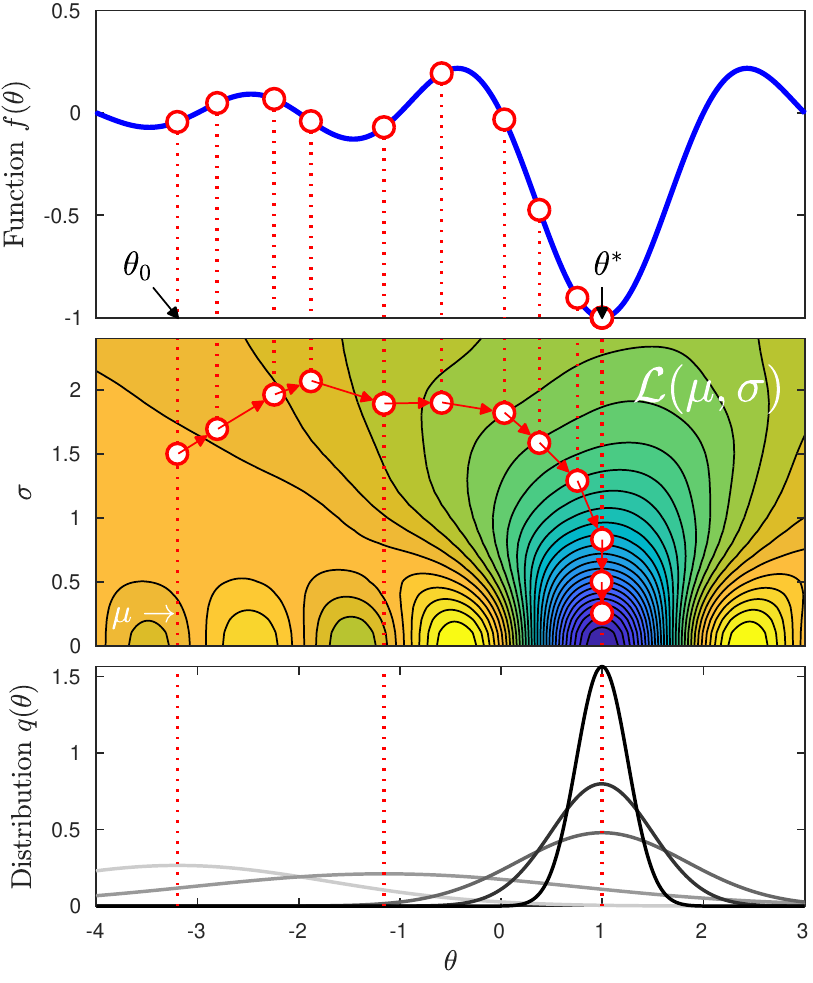}
\caption{Illustrative application of VAN on an example by \cite{ferencBlogES}. The top figure shows the function $f(\theta) =\textrm{sinc}(\theta)$ with a blue curve. The global minima is at $\vtheta^*=1$ although there are many local minima and maxima as well. The second plot shows the VO objective $\mathcal{L}(\mu,\sigma) = \myexpect_q[f(\theta)]$ for the Gaussian $q = \gauss(\theta|\mu,\sigma^2)$. The red points and arrows show the iterations of our VAN method initialized at $\mu=-3.2$ and $\sigma=1.5$.
   %The markers are shown at iterations $t = 1, 10, 15, 17, 20, 22, 25, 27, 30, 35, 40$ and $50$.
   The $\theta$ values corresponding to the value of $\mu$ are marked in the top figures where we see that these iterations converge to the global minima of $f$ and avoids other local optima. When we use a Newton's method initialized at the same point, i.e., $\theta_0 = -3.2$, it converges to the local minima at -3.5. VAN can avoid such local minima because it optimizes in the space of $\mu$ and $\sigma^2$. The progression of the distribution $q$ is shown in the bottom figure, where darker curves indicate higher
iterations. We see that the distribution $q$ is flat in the beginning which enables more exploration which in turn helps the method to avoid local minima. As desired, the distribution peaks around $\theta^*$ as iterations increase.}
\label{fig:van_illustration}
\end{figure}

Our main contribution is a new method to solve \eqref{eq:VO} by using a mirror-descent algorithm. We show that our algorithm is a second-order method which solves the original problem \eqref{eq:minf}, even though it is designed to solve the problem \eqref{eq:VO}. Due to its similarity to Newton's method, we refer to our method as the Variational Adaptive Newton (VAN) method.
Figure \ref{fig:van_illustration} shows an example of our method for a one-dimensional non-convex function.

We establish connections of our method to many existing methods in continuous optimization, variational inference, and evolution strategies, and use these connections to derive new algorithms for explorative learning. Below, we summarize the contributions made in the rest of the paper:
\begin{itemize}
   \item In Section \ref{sec:van}, we derive VAN and establish it as a second-order method. In Section \ref{sec:large}, we derive computationally-efficient versions of VAN and discuss their relations to adaptive-gradient methods.
   \item In Section \ref{sec:vi} and \ref{sec:nes}, we show connections to variational inference methods and natural evolution strategy \citep{wierstra2008natural}.  
\item In Section \ref{sec:experiments}, we apply our method to supervised leaning, unsupervised learning, active learning, and reinforcement learning. In Section \ref{sec:discussion}, we discuss relevance and limitations of our approach.
\end{itemize}
%\comment{Needs work}
%Our work exploits the power of explorative learning to solve difficult learning a wide variety of learning tasks, while requiring computations similar to continuous optimization methods.
This work presents a general-purpose explorative-learning method that has the potential to improve learning in areas such as active learning and reinforcement learning.

\section{Variational Optimization}
\label{sec:vo}
We will focus on solving the problem \eqref{eq:VO} since it enables estimation of a distribution that can be used for exploration. In this paper, we will use the Gaussian distribution $q(\vtheta)$. The problem \eqref{eq:VO} can then be rewritten as follows,
\begin{align}
   \min_{\mu,\Sigma} \myexpect_{\mathcal{N}(\theta|\mu,\Sigma)} \sqr{f(\vtheta)} := \mathcal{L}(\vmu,\vSigma) 
   \label{eq:minEf}
\end{align}
where $q$ is the Gaussian distribution $q(\vtheta|\veta) := \gauss(\vtheta|\vmu,\vSigma)$ with $\vmu$ being the mean and $\vSigma$ being the covariance, and $\veta = \{\vmu,\vSigma\}$.
The function $\mathcal{L}$ is differentiable under mild conditions even when $f$ is not differentiable, as discussed in \cite{2012arXiv1212.4507S}. This makes it possible to apply gradient-based optimization methods to optimize it.
%The parameter $\veta$ usually consist of either the mean and covariance, but we can also choose other parameterizations, e.g., natural parameterization $\veta = \{\vSigma^{-1}\vmu, -\half\vSigma^{-1}\}$. It is also possible to use other distribution as well, but the Gaussian is the most popular choice and has been shown to work well in practice on a wide variety of problems.

A straightforward approach to minimize $\mathcal{L}$ is to use Stochastic Gradient Descent (SGD) as shown below,
\begin{align}
%\veta_{t+1} = \veta_t - \rho_t \left[ \widehat{\nabla}_{\eta} \myexpect_q\sqr{f(\vtheta)} \right]_{\eta=\eta_t}
   \textrm{V-SGD }: \quad \vmu_{t+1} &= \vmu_t - \rho_t \sqr{\sg{\mathcal{L}_t}{\mu} \label{eq:vsgd_mean}}\\
   \vSigma_{t+1} &= \vSigma_t - \rho_t \sqr{ \sg{\mathcal{L}_t}{\Sigma} } \label{eq:vsgd_cov}
\end{align}
where $\rho_t>0$ is a step size at iteration $t$, $\widehat{\nabla}$ denotes an unbiased stochastic-gradient estimate, and $\mathcal{L}_t = \mathcal{L}(\bmu_t, \bSigma_t)$. We refer to this approach as Variational SGD or simply V-SGD to differentiate it from the standard SGD that optimizes $f(\vtheta)$ in the $\vtheta$ space.

The V-SGD approach is simple and can also work well when used with adaptive-gradient methods to adapt the step-size, e.g., AdaGrad and RMSprop.
However, as pointed by \cite{wierstra2008natural}, it has issues, especially when REINFORCE \citep{williams1992simple} is used to estimate the gradients of $f(\btheta)$. \cite{wierstra2008natural} argue that the V-SGD update becomes increasingly unstable when the covariance is small, while becoming very small when the covariance is large.
To fix these problems, \citet{wierstra2008natural} proposed a natural-gradient method.
Our method is also a natural-gradient method, but, as we show in the next section, its updates are much simpler and they lead to a second-order method which is similar to Newton's method.
%(this is discussed in detail in Section \ref{}).

\section{Variational Adaptive-Newton Method}
\label{sec:van}
VAN is a natural-gradient method derived using a mirror-descent algorithm. Due to this, the updates of VAN are fundamentally different from V-SGD. We will show that VAN adapts the step-sizes in a very similar spirit to the adaptive-gradient methods. This property will be crucial in establishing connections to Newton's method.

VAN can be derived by making two minor modifications to the V-SGD objective. 
Note that the V-SGD update in \eqref{eq:vsgd_mean} and \eqref{eq:vsgd_cov} are solutions of following optimization problem:
\begin{align}
\veta_{t+1} = \argmin_{\boldsymbol{\eta}} \,\, \left\langle \veta , \sg{\mathcal{L}_t}{\eta} \right\rangle  + \frac{1}{2\rho_t} \| \veta-\veta_t \|^2 .
\label{eq:v-sgd-1}
\end{align}
The equivalence to the update \eqref{eq:vsgd_mean} and \eqref{eq:vsgd_cov} can be shown by simply taking the derivative with respect to $\veta$ of \eqref{eq:v-sgd-1} and setting it to zero:
\begin{align}
\sg{\mathcal{L}_t}{\eta}  + \frac{1}{\rho_t} (\veta_{t+1}-\veta_t) = 0 ,
\end{align}
which results in the update \eqref{eq:vsgd_mean} and \eqref{eq:vsgd_cov}.
A simple interpretation of this optimization problem is that, in V-SGD, we choose the next point $\veta$ along the gradient but contain it  within a scaled $\ell_2$-ball centered at the current point $\veta_t$.
This interpretation enables us to obtain VAN by making two minor modifications to V-SGD.

The first modification is to replace the Euclidean distance $\| \cdot \|^2$ by a \emph{Bregman} divergence which results in the \emph{mirror-descent} method.
Note that, for exponential-family distributions, the \emph{Kullback-Leibler} (KL) divergence corresponds to the Bregman divergence \citep{raskutti2015information}. Using the KL divergence results in natural-gradient updates which results in better steps when optimizing the parameter of a probability distribution~\citep{amari1998natural}.

The second modification is to optimize the VO objective with respect to the mean parameterization of the Gaussian distribution $\vm := \{\vmu,\vmu\vmu^T + \vSigma\}$ instead of the parameter $\veta := \{\vmu, \vSigma\}$.
We emphasize that this modification does not change the solution of the optimization problem since the Gaussian distribution is a \emph{minimal} exponential family and the relationship between $\veta$ and $\vm$ is one-to-one (see Section 3.2 and 3.4.1 of \cite{WainwrightJordan08} on the basics of exponential family and mean-parameterization respectively).

The two modifications give us the following problem:
\begin{align}
   \vm_{t+1} = \argmin_{\boldsymbol{m}}\,\, \left\langle \vm, \sg{\mathcal{L}_t}{m} \right\rangle + \frac{1}{\beta_t} \dkls{}{q}{q_t}, \label{eq:md}
\end{align}
where $q:=q(\vtheta|\vm)$, $q_t:=q(\vtheta|\vm_t)$, and $\dkls{}{q}{q_t} = \mathbb{E}_{q}[\log(q/q_t)]$ denotes the KL divergence. The convergence of this procedure is guaranteed under mild conditions \citep{ghadimi2014mini}.

As shown in Appendix~\ref{appendix:derivation_van}, a solution to this optimization problem is given by
\begin{align}
\vmu_{t+1} &= \vmu_{t} - \beta_t \,\, \vSigma_{t+1} \sqr{\sg{\mathcal{L}_t}{\mu}}  \label{eq:Van_mean_0}\\
\vSigma_{t+1}^{-1} &= \vSigma_t^{-1} + \,\, 2\beta_t \,\, \sqr{ \sg{\mathcal{L}_t}{\Sigma}}  \label{eq:Van_prec_0}
\end{align}
The above update differs from those of V-SGD in two ways.
First, here we update the precision matrix $\vSigma^{-1}$ while in V-SGD we update the covariance matrix $\vSigma$. However, both updates use the gradient with respect to $\vSigma$.
Second, the step-size for $\vmu_{t+1}$ are \emph{adaptive} in the above update since $\beta_t$ is scaled by the covariance $\vSigma_{t+1}$.  

%In the next section, we describe an important property of VAN and show that VAN is a second-order optimization method.
%Then, in the following sections, we draw connections between VAN and existing methods for variational inference and evolution strategies. 
%
%
%\subsection{VAN for Continuous Optimization}
%\section{Explorative Learning via VAN}
%\section{VAN as a Second-order Method}
%\label{sec:second}
The above updates corresponds to a second-order method which is very similar to Newton's method.
We can show this using the following identities
\citep{Opper:09}:
\begin{align}
\nabla_{\mu} \myexpect_q \sqr{f(\vtheta) } &= \myexpect_q \sqr{ \nabla_\theta f(\vtheta)} 
%\approx \frac{1}{S} \sum_{s=1}^S \nabla_\theta f(\vtheta^{(s)}) 
, \label{eq:bonnet}\\
\nabla_\Sigma \myexpect_q \sqr{f(\vtheta) } &= \half \myexpect_q \sqr{ \nabla^2_{\theta\theta} f(\vtheta) } . 
%\approx \frac{1}{2S} \sum_{s=1}^S  \nabla^2_{\theta\theta} f(\vtheta^{(s)}),  \nonumber
\end{align}
By substituting these into \eqref{eq:Van_mean_0} and \eqref{eq:Van_prec_0}, we get the following updates which we call the VAN method:
\begin{align}
   \textrm{VAN: } \quad \vmu_{t+1} &= \vmu_{t} - \beta_t \,\, \vP_{t+1}^{-1}  \myexpect_{q_t} \sqr{  \widehat{\nabla}_\theta f(\vtheta)} \label{eq:Van_mean_1}\\
   \vP_{t+1} &= \vP_t + \,\, \beta_t \,\, \myexpect_{q_t} \sqr{ \widehat{\nabla}^2_{\theta\theta} f(\vtheta)} \label{eq:Van_prec_1}
\end{align}
where $\vP_t := \vSigma_t^{-1}$ is the precision matrix and $q_t := \gauss(\vtheta|\vmu_t, \vSigma_t)$.
The precision matrix $\vP_t$ contains a running-sum of the past \emph{averaged Hessians}, and the search-direction for the mean is obtained by scaling the \emph{averaged gradients} by the inverse of $\vP_t$.
If we compare Eq. \ref{eq:Van_mean_0} to the following update of Newton's method,
\begin{align}
\vtheta_{t+1} &= \vtheta_t - \rho_t \sqr{\nabla^2_{\theta\theta} f(\vtheta_t)}^{-1} \sqr{\nabla_\theta f(\vtheta_t)},
\end{align}
we can see that the Hessian matrix is replaced by the Precision matrix in the VAN update.
%where the search direction is obtained by scaling the gradients at $\btheta_t$ with the inverse Hessian at $\btheta_t$.
Due to this connection to Newton's method and the use of an adaptive scaling matrix $\vP_t$, we call our method the \emph{Variational Adaptive-Newton} (VAN) method.

%The running-sum makes the scaling matrix $\vP_t$ adaptive in VAN.
%For this reason and due to its similarity to Newton's method, we call our method the Variational Adaptive-Newton (VAN) method.

The averaged gradient and running sum of averaged Hessian allow VAN to avoid some types of local optima.
Figure~\ref{fig:van_illustration} shows such a result when minimizing\footnote{This example is discussed in a blog by \cite{ferencBlogES}} $f(\theta) = \mathrm{sinc}(\theta)$ with initial solution at $\theta = -3.2$.
Clearly, both the gradient and Hessian at $\theta_0$ suggest updating $\theta$ towards a local optimum at $\theta = -3.5$.
However, VAN computes an averaged gradient over samples from $q(\theta)$ which yields steeper descent directions pointing towards the global optimum. The adaptive scaling of the steps further ensures a smooth convergence.

The averaging property of VAN is strikingly different from other second-order optimization methods. We expect VAN to be more robust due to averaging of gradients and Hessians.
Averaging is particularly useful for optimization of stochastic objectives. For such objectives, application of Newton's method is difficult because reliably selecting a step-size is difficult. Several variants of Newton's method have been proposed to solve this difficulty, e.g., based on quasi-Newton methods \citep{byrd2016stochastic} or incremental approaches \citep{gurbuzbalaban2015globally}, or by simply adapting mini-batch size \citep{mokhtari2016adaptive}.
VAN is most similar to the incremental approach of \citep{gurbuzbalaban2015globally} where a running sum of past Hessians is used instead of just a single Hessian. In VAN however the Hessian is replaced by the average of Hessians with respect to $q$. 
For stochastic objectives, VAN differs substantially from existing approaches and it has the potential to be a viable alternative to them.

%The robustness of VAN also makes it more suitable to stochastic learning setting when compared with Newton method since stochasticity makes it to choose an appropriate step-size. Several stochastic versions of Newton's method have been proposed to solve this problem, e.g. quasi-Newton methods \cite{byrd2016stochastic}, incremental approach \cite{gurbuzbalaban2015globally}, adapting mini-batch size to select step-size \cite{mokhtari2016adaptive}.
% \cite{shamir2014communication} uses a mirror-descent approach that enables distributed computation.   
%VAN is most similar to the incremental approach \cite{gurbuzbalaban2015globally} and could be a viable second-order alternative for stochastic objectives.

%VAN is suitable for stochastic settings, e.g., when $f$ corresponds to the log-likelihood with large number of data points. 
%The stochastic noise in such cases makes it challenging to use Newton method because step-size selection is difficult. Several stochastic versions of Newton's method have been proposed to solve this problem, e.g. quasi-Newton methods \cite{byrd2016stochastic}, incremental approach \cite{gurbuzbalaban2015globally}, adapting mini-batch size to select step-size \cite{mokhtari2016adaptive}.
% \cite{shamir2014communication} uses a mirror-descent approach that enables distributed computation.   
%VAN is most similar to the incremental approach \cite{gurbuzbalaban2015globally} and could be a viable second-order alternative for stochastic objectives.

An issue with using the Hessian is that it is not always positive semi-definite, for example, for non-convex problems. For such cases, we can use a Gauss-Newton variant shown below \cite{bertsekas1999nonlinear} which we call the Variational Adaptive Gauss-Newton (VAG) Method:
\begin{align}
   \textrm{VAG: } \vP_{t+1} &= \vP_t + \beta_t \myexpect_{q_t} \sqr{ \sqr{\nabla_{\theta} f(\vtheta) } \sqr{\nabla_{\theta} f(\vtheta)}^T} .  \label{eq:Vag_prec}
   \end{align}

\section{VAN for Large-Scale Problems}
\label{sec:large}

Applying VAN to problems with large number of parameters is not feasible because we cannot compute the exact Hessian matrix. In this section, we describe several variants of VAN that scale to large problems. Our variants are similar to existing adaptive-gradient methods such as AdaGrad~\cite{duchi2011adaptive} and AROW~\citep{crammer2009adaptive}. We derive these variants by using a mean-field approximation for $q$. Our derivation opens up the possibility of a new framework for designing
computationally efficient second-order methods by using structured distributions $q$.

One of the most common way to obtain scalability is to use a diagonal approximation of the Hessian. In our case, this approximation corresponds to a distribution $q$ with diagonal covariance, i.e., $q(\vtheta|\veta) = \prod_{d=1}^D \gauss(\theta_d|\mu_d,\sigma_d^2)$, where $\sigma_d^2$ is the variance.
 This is a common approximation in variational inference methods and is called the mean-field approximation \citep{bishop2006pattern}.
Let us denote the precision parameters by $s_d = 1/\sigma_d^2$, and a vector containing them by $\vs$. Using this Gaussian distribution in the update \eqref{eq:Van_mean_1} and \eqref{eq:Van_prec_1}, we get the following diagonal version of VAN, which we call VAN-D: 
   \begin{align}
      \textrm{ VAN-D: } \vmu_{t+1} &= \vmu_{t} - \beta_t \,\, \diag(\vs_{t+1})^{-1} \myexpect_{q_t} \sqr{  \nabla_\theta f(\vtheta)}  \nonumber\\
      \vs_{t+1} &= \vs_t + \,\, \beta_t \,\, \myexpect_{q_t} \sqr{\vh(\vtheta)}  \label{eq:Vand_prec_0}
   \end{align}
   where $\diag(\vs)$ is a diagonal matrix containing the vector $\vs$ as its diagonal and $\vh(\vtheta)$ is the diagonal of the Hessian $\nabla^2_{\theta\theta} f(\vtheta)$.
%A strategy similar to this is applied for natural-gradients in deep neural networks by \cite{grosse2015scaling, martens2015optimizing}, although they consider an approximation on the model instead of a distribution. 
%In our framework, we can design scalable algorithms by restricting the structure of $q$ without making any approximation about the model. We expect this property to be useful.

The VAN-D update requires computation of the expectation of the diagonal of the Hessian, which could still be difficult to compute. Fortunately, we can compute its approximation easily by using the reparameterization trick \citep{kingma2014adam}. This is possible in our framework because we can express the expectation of the Hessian as gradients of an expectation, as shown below:
\begin{align}
\myexpect_q\sqr{\nabla_{\theta_d \theta_d}^2 f(\vtheta) } &= 2 \nabla_{\sigma_d^2} \sqr{\myexpect_q[f(\vtheta)] } , \\
                                                          & = 2 \nabla_{\sigma_d^2} \myexpect_{\mathcal{N}(\epsilon|0,1)} \sqr{ f(\mu_d + \sigma_d \epsilon) } ,\\
& = 2 \myexpect_{\mathcal{N}(\epsilon|0,1)}\sqr{  \nabla_{\sigma_d^2} f(\mu_d + \sigma_d \epsilon) } , \label{eq:reparam}
\end{align}
where the first step is obtained using \eqref{eq:bonnet}.
In general, we can use the stochastic approximation by simultaneous perturbation (SPSS) method \citep{bhatnagar2007adaptive, spall2000adaptive} to compute derivatives. A recent paper by \cite{2017arXiv170303864S} showed that this type of computation can also be parallelized which is extremely useful for large-scale learning.
Note that these tricks cannot be applied directly to standard continuous-optimization methods. The presence of expectation with respect to $q$ in VO enables us to leverage such stochastic approximation methods for large-scale learning.

Finally, when $f(\vtheta)$ corresponds to a supervised or unsupervised learning problem with large data, we could compute its gradients by using stochastic methods. We use this version of VAN in our large-scale experiments and call it sVAN, which is an abbreviation for stochastic-VAN.

\subsection{Relationship with Adaptive-Gradient Methods}
The VAN-D update given in \eqref{eq:Vand_prec_0} is closely-related to an adaptive-gradient method called AdaGrad \citep{duchi2011adaptive} which uses the following updates:
\begin{align}
 \textrm{AdaGrad: } \vtheta_{t+1} &= \vtheta_t - \rho_t \diag(\vs_{t+1})^{-1/2} \vg(\vtheta_t) \label{eq:adgrad2} \\
 \vs_{t+1} &= \vs_t + [\vg(\vtheta_t)\odot\vg(\vtheta_t)] 
\end{align}
Comparing these updates to \eqref{eq:Vand_prec_0}, we can see that both AdaGrad and VAN-D compute the scaling vector $\vs_t$ using a moving average.
However, there are some major differences between their updates: 1. VAN-D uses average gradients instead of a gradient at a point, 2. VAN-D does not raise the scaling matrix to the power of 1/2, 3. The update of $\vs_t$ in VAN-D uses the diagonal of the Hessian instead of the squared gradient values used in AdaGrad. It is possible to use the squared gradient in VAN-D but since we can compute the Hessian using
the reparameterization trick, we do not have to make further approximations. 

VAN can be seen as a generalization of an adaptive method called AROW~\citep{crammer2009adaptive}. AROW uses a mirror descent algorithm that is very similar to ours, but has been applied only to problems which use the Hinge Loss. Our approach not only generalizes AROW but also brings new insights connecting ideas from many different fields.

\begin{table}[!t]
\centering
\caption{Variants of VAN used in our experiments.}
\label{tab:listofmethods}
\vspace{.1in}
\resizebox{0.95\columnwidth}{!} {
\begin{tabular}{ |l| l| }
  \hline
  \textbf{Method} & \textbf{Description} \\
  \hline            
  VAN & Variational Adaptive-Newton Method using \eqref{eq:Van_mean_1} and \eqref{eq:Van_prec_1}.\\
  sVAN & Stochastic VAN (gradient estimated using mini-batches). \\
  sVAN-Exact & Stochastic VAN with no MC sampling. \\
  sVAN-D & Stochastic VAN with diagonal covariance $\Sigma$. \\
  sVAN-Active & Stochastic VAN using active learning. \\
  sVAG & Stochastic Variational Adaptive Gauss-Newton Method.\\
  sVAG-D & Stochastic VAG with diagonal covariance $\Sigma$. \\
  \hline
\end{tabular}
}
\end{table}

\section{VAN for Variational Inference}
\label{sec:vi}
Variational inference (VI) enables a scalable computation of the Bayesian posterior and therefore can also be used for explorative learning. In fact, VI is closely related to VO. In VI, we compute the posterior approximation $q(\vtheta|\veta)$ for a model $p(\vy,\vtheta)$ with data $\vy$ by minimizing the following objective: 
\begin{align}
   \min_{\boldsymbol{\eta}} \myexpect_{q(\boldsymbol{\theta}|\boldsymbol{\eta})} \sqr{ - \log \frac{p(\vy,\vtheta)}{ q(\vtheta|\veta) }} := \mathcal{L}_{VI}(\veta) \label{eq:vi}
\end{align}
We can perform VI by using VO type of method on the function inside the square bracket. A small difference here is that the function to be optimized also depends on parameters $\veta$ of $q$.
Conjugate-computation variational inference (CVI) is a recent approach for VI by \cite{khan2017conjugate}. We note that by applying VAN to the variational objective, one recovers CVI. VAN however is more general than CVI since it applies to many other problems other than VI. A direct consequence of our connection is that CVI, just like VAN, is also a second-order method to optimize $f_{VI}(\vtheta)$, and is related to adaptive-gradient methods as well. Therefore, using CVI
should give better results than standard methods that use update similar to V-SGD, e.g., black-box VI method of \cite{ranganath2013black}.

\section{VAN for Evolution Strategies}
\label{sec:nes}
VAN performs natural-gradient update in the parameter-space of the distribution $q$ (as discussed earlier in Section \ref{sec:van}). The connection to natural-gradients is based on a recent result by \citet{raskutti2015information} that shows that the mirror descent using a KL divergence for exponential-family distributions is equivalent to a natural-gradient descent. The natural-gradient corresponds to the one obtained using the Fisher information of the exponential family distribution. In our case, the mirror-descent algorithm \eqref{eq:md} uses the Bregman divergence that corresponds to the KL divergence between two Gaussians. Since the Gaussian is an exponential-family distribution, mirror descent
\eqref{eq:md} is equivalent to natural-gradient descent in the dual Riemannian manifold of a Gaussian. Therefore, VAN takes a natural-gradient step by using a mirror-descent step.

Natural Evolution Strategies (NES) \citep{wierstra2008natural} is also a natural-gradient algorithm to solve the VO problem in the context of evolution strategies. NES directly applies natural-gradient descent to optimize for $\bmu$ and $\bSigma$ and this yields an infeasible algorithm since the  Fisher information matrix has $O(D^4)$ parameters.
To overcome this issue, \cite{wierstra2008natural} proposed a sophisticated re-parameterization that reduces the number of parameters to $O(D^2)$.
VAN, like NES, also has $O(D^2)$ parameters, but with much simpler updates rules due to the use of mirror descent in the mean-parameter space.

\section{Applications and Experimental Results}
\label{sec:experiments}
In this section, we apply VAN to a variety of learning tasks to establish it as a general-purpose learning tool, and also to show that it performs comparable to continuous optimization algorithm while extending the scope of their application. Our first application is supervised learning with Lasso regression. Standard second-order methods such as Newton's method cannot directly be applied to such problems because of discontinuity.
For this problem, we show that VAN enables \emph{stochastic} second-order optimization which is faster than existing second-order methods such as iterative-Ridge regression.
We also apply VAN to supervised learning with logistic regression and unsupervised learning with Variational Auto-Encoder, and show that stochastic VAN gives comparable results to existing methods such as AdaGrad. Finally, we show two application of VAN for explorative learning, namely active learning for logistic regression and parameter-space exploration for deep reinforcement learning.

Table \ref{tab:listofmethods} summarizes various versions of VAN compared in our experiments. The first method is the VAN method which implements the update shown in \eqref{eq:Van_mean_1} and \eqref{eq:Van_prec_1}. Stochastic VAN implies that the gradients in the updates are estimated by using minibatches of data. The suffix `Exact' indicates that the expectations with respect to $q$ are computed exactly, i.e., without resorting to Monte Carlo (MC) sampling. This is possible for
the Lasso objective
as shown in \cite{2012arXiv1212.4507S} and
also for logistic regression as shown in \cite{marlin2011piecewise}. The suffix `D' indicates that a diagonal covariance with the update \eqref{eq:Vand_prec_0} is used.
The suffix `Active' indicates that minibatches are selected using an active learning method.
Final, VAG corresponds to the Gauss-Newton type update shown in \eqref{eq:Vag_prec}.
For all methods, except `sVAN-Exact', we use MC sampling to approximate the expectation with respect to $q$. In our plot, we indicate the number of samples by adding it as a suffix, e.g., sVAN-10 is the stochastic VAN method with 10 MC samples.

%In this section, we demonstrate the usefulness of VAN on the following machine learning tasks: regression, classification, active learning for classification, variational auto-encoder, and reinforcement learning. The list of proposed methods used in our experiments and their abbreviations are given in Table \ref{tab:listofmethods}. The numbers in abbrevtiations indicates number of samples used to approximate the averaged-gradient and Hessian (when it is not available in
%closed form).

\subsection{Supervised Learning: Lasso Regression}

%The goal of regression is to learn a predictive model from observations $\{(\bx_i, y_i) \}_{i=1}^N \sim p(\bx, y)$ such that it accurately predicts the output $y$ of unseen data.
%We assume a linear predictive model $\btheta^\top \bx$ where $\btheta \in \mathbb{R}^D$ is a parameter vector to be learned. 
Given $N$ example pairs $\{y_i, \vx_i\}_{i=1}^N$ with $\vx_i\in\real^D$, in lasso regression, we minimize the following loss that contains an $\ell_1$-regularization:
\begin{align}
   f(\vtheta) = \sum_{i=1}^{N} (y_i - \vtheta^T \bx_i )^2 + \lambda \sum_{d=1}^D | \theta_d |,
%f(\vtheta) = \sum_{i=1}^{N} (y_i - \vtheta^T \bx_i )^2 + \lambda \sum_{d}^{D} |\theta_d|.
\end{align}
Because the function is non-differentiable, we cannot directly apply gradient-based methods to solve the problem. For the same reason, it is also not possible to use second-order methods such as Newton's method. VAN can be applied to this method since expectation of $|\theta_d|$ is twice differentiable. We use the gradient and Hessian expression given in \cite{2012arXiv1212.4507S}.

We compare VAN and sVAN with the iterative-ridge method (iRidge), which is also a second-order method. We compare on two datasets: Bank32nh ($N=8192, D=256, N_{train} = 7290, \lambda = 104.81$) and YearPredictionMSD ($N=515345, D=90, N_{train} = 448350, \lambda = 5994.84$). 
We set $\lambda$ values using a validation set where we picked the value that gives minimum error over multiple values of $\lambda$ on a grid.
The iRidge implementation is based on {\tt minFunc} implementation by Mark Schmidt. 
For sVAN, the size of the mini-batch used to train Bank32nh and YearPredictionMSD are $M=30$ and $M=122$ respectively.
%Dataset statistics and model hyper-parameters are available in the appendix \ref{appendix:datasets}. 
We report the absolute difference of parameters, $\vtheta-\vtheta^*$ where $\vtheta$ is the parameters estimated by a method and $\vtheta^*$ is the parameters optimal value (found by iRidge). For VAN the estimated value is equal to the mean $\vmu$ of the distribution.
Results are shown in Figure \ref{figure:mix} (a) and (b), where we observe that VAN and iRidge perform comparably, but sVAN is more data-efficient than them in the first few passes. Results on multiple runs show very similar trends.

In conclusion, VAN enables application of a stochastic second-order method to a non-differentiable problem where existing second-order method and their stochastic versions cannot be applied directly.

\subsection{Supervised Learning: Logistic Regression}

%The goal of classification is to learn a classifier that accurately predict a class label $c$ of an input $\bx$ using a labeled dataset $\{(\bx_i, c_i) \}_{i=1}^N$.
%Classification is another common machine learning task.
%We consider a probabilistic binary classification task where an input sample $\bx$ belongs to one of the two classes $y \in \{-1, 1\}$ and the goal is to learn a posterior distribution $p(y=-1|\bx)$ and $p(y=1|\bx)$ from a training dataset $\mathcal{D} = \{(\bx_i, y_i) \}_{i=1}^N \sim p(\bx, y)$.
%A common approach is logistic regression which aims to learn a parameter vector $\btheta \in \mathbb{R}^{D}$ of a linear model with the logistic activation function: $\widehat{p}(y|\bx , \btheta) = \sigma(y \cdot  \btheta^\top\bx)$, 
%where $\sigma(a) = \frac{1}{1+e^{-a}}$.
In logistic regression, we minimize the following:
\begin{align}
f(\btheta) 
%&= -\mathbb{E}_{p(\bx,y)}\left[ \log \widehat{p}(y|\bx, \btheta)\right] + \lambda\|\btheta\|_2^2 \notag \\
%&= -\mathbb{E}_{p(\bx,y)}\left[ \log 1 - \log(1+e^{-y \cdot \btheta^\top\bx}) \right] \notag \\
&= \sum_{i=1}^N \left[ \log(1+e^{y_i (\btheta^\top\bx_i)}) \right] + \lambda \sum_{d=1}^D \theta_d^2 ,
\end{align}
where $y_i \in\{-1,+1\}$ is the label.
%Note that this is a convex function and all learning methods obtain the same optimal solution.
%We consider two settings for minimizing $f(\btheta)$.
%The first is batch learning where in each learning iteration the expectation over $p(\bx,y)$ is approximated using the entire training dataset $\mathcal{D}$.
%The second is stochastic learning where in each learning iteration the expectation over $p(\bx,y)$ is approximated using a mini-batch $\{(\bx_j,y_j)\}_{j=1}^M$ drawn uniformly from $\mathcal{D}$ where $M \ll N$.

We compare VAN to Newton's method and AdaGrad both of which standard algorithms for batch and stochastic learning, respectively, on convex problems. We use VAN, its stochastic version sVAN, and the diagonal version sVAN-D.
We use three real-world datasets from the libSVM database~\citep{CC01a}: `breast-cancer-scale' ($N=683,D=10,N_{train}=341,\lambda=1.88$), `USPS' ($N=1,540,D=256,N_{train}=770,\lambda=6.21$), and `a1a' ($N=32,561,D=123,N_{train}=1,605,\lambda=67.23$).
We compare the log-loss on the test set computed as follows: $ \sum_i \log (1+ \exp(y_i (\hat{\vtheta}^T\vx_i))/N_{test}$ where $\hat{\vtheta}$ is the parameter estimate and $N_{test}$ is the number of examples in the test set.
For sVAN and sVAN-D, we use a mini-batch size of 10 for `breast-cancer-scale' dataset and a mini-batch of size 100 for the rest of the datasets.

\begin{figure*}[t]
	\centering
	\includegraphics[width=0.33\linewidth]{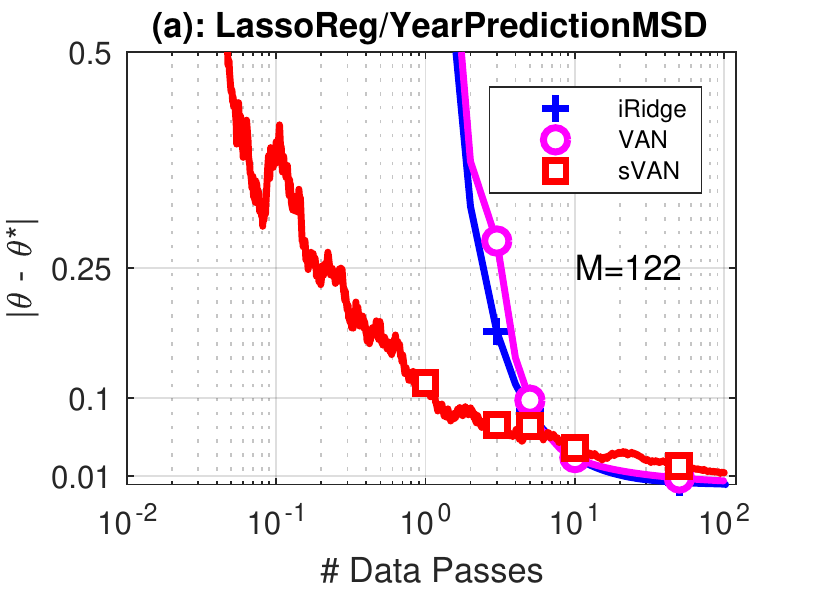}
	\includegraphics[width=0.33\linewidth]{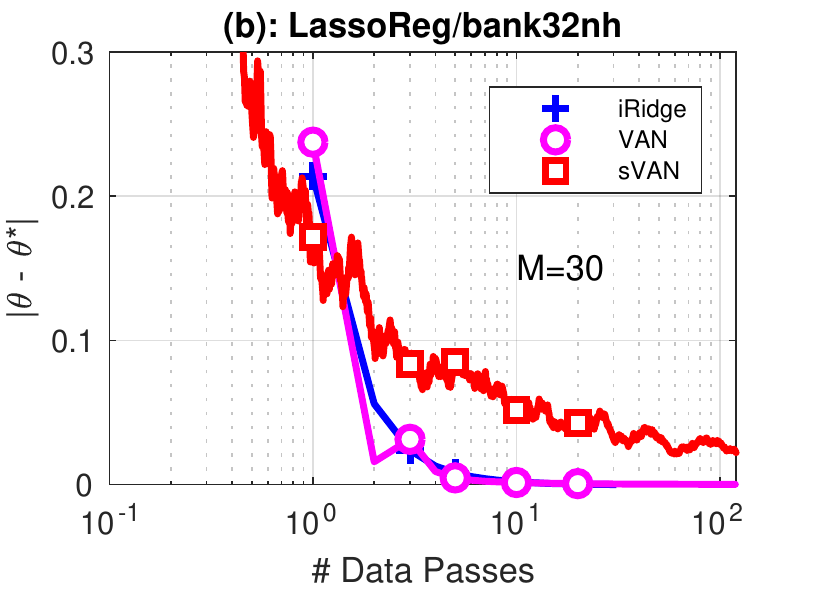}
	\includegraphics[width=0.33\linewidth]{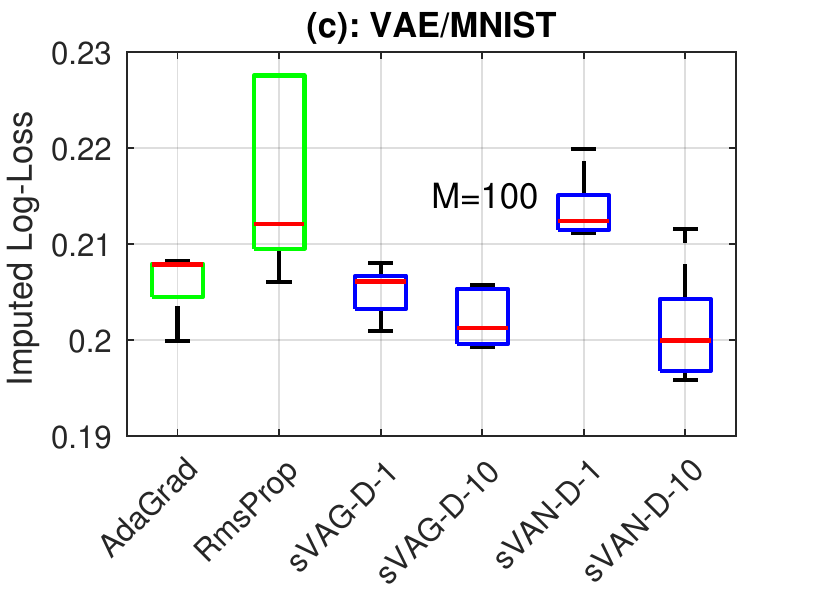}
	\includegraphics[width=0.33\linewidth]{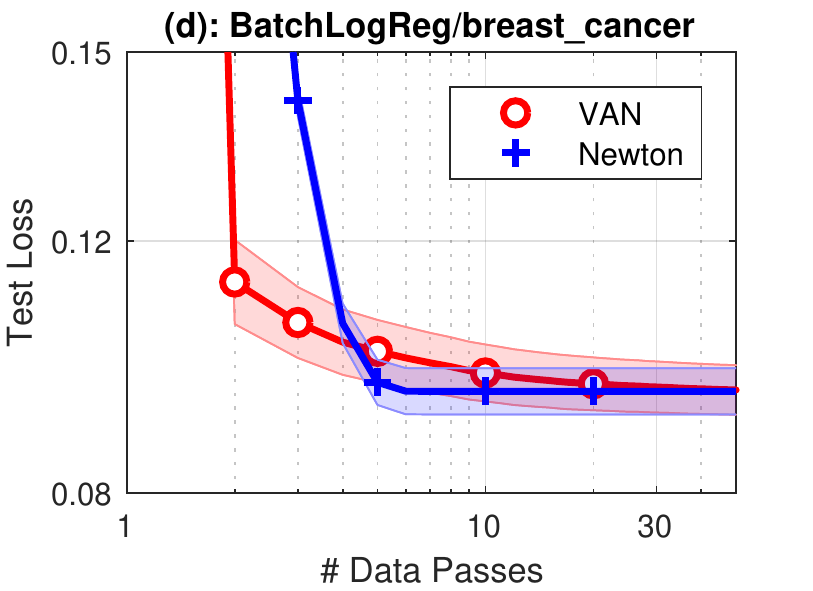}
	\includegraphics[width=0.33\linewidth]{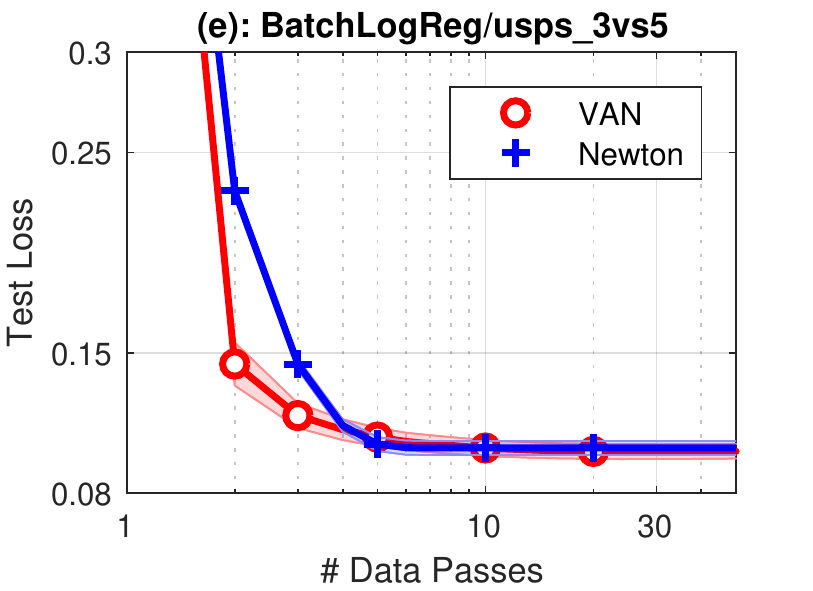}
	\includegraphics[width=0.33\linewidth]{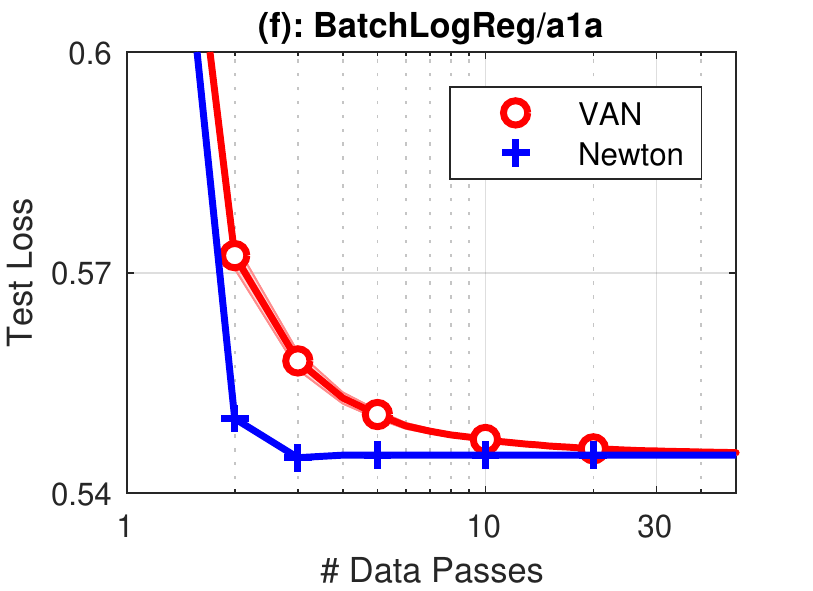}
	\includegraphics[width=0.33\linewidth]{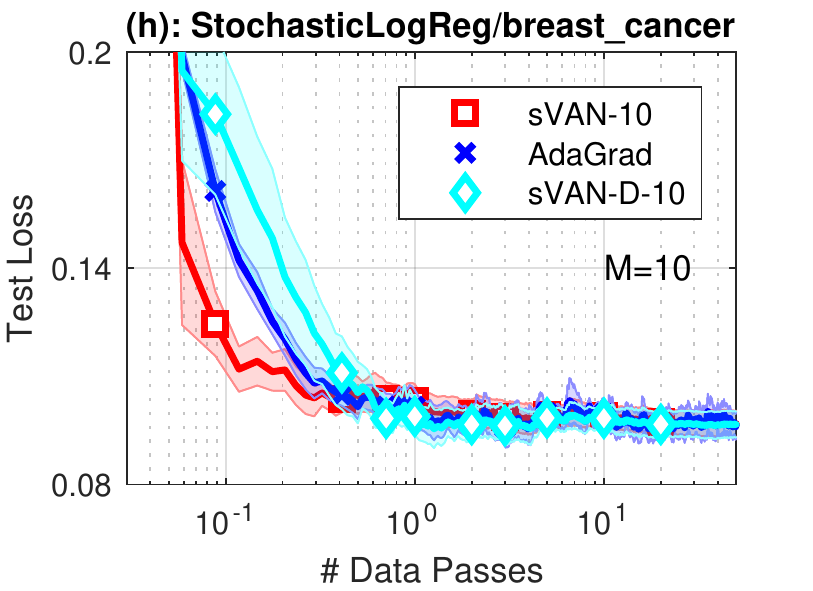}
	\includegraphics[width=0.33\linewidth]{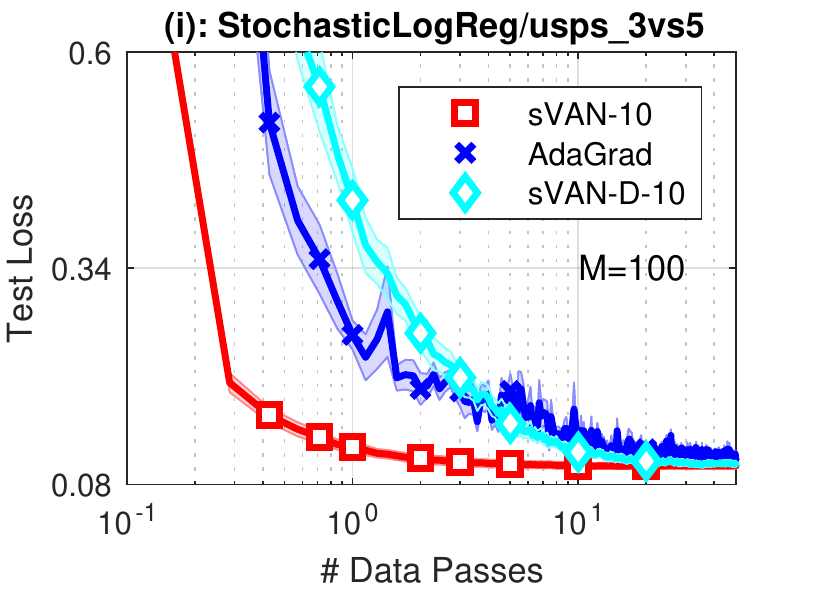}
	\includegraphics[width=0.33\linewidth]{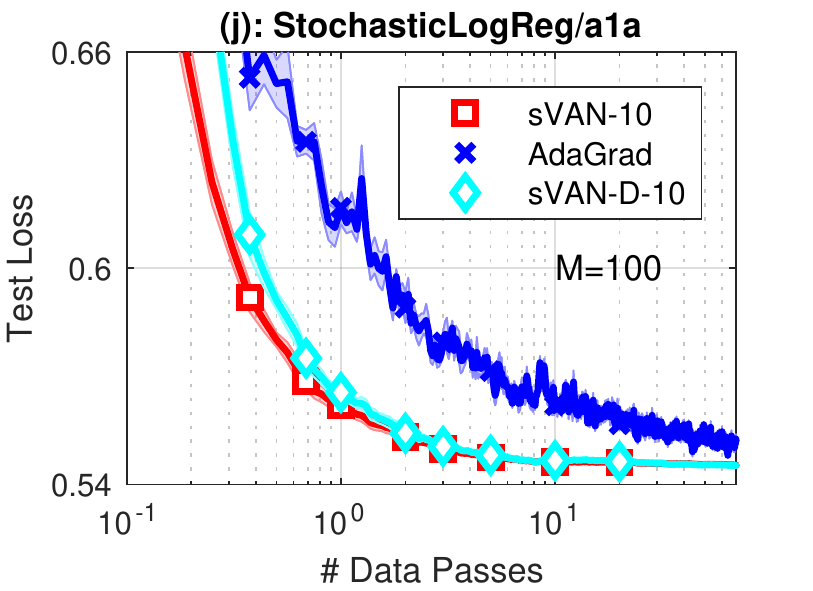}
	\caption{Experimental results on different learning tasks. (a-b): Lasso regression. (c): VAE. (d-f):  Logistic regression with VAN and Newton's method. (h-j): Logistic regression with sVAN, sVAN-D and AdaGrad. Datasets are specified in the title. $M$ refers to the mini-batch size for stochastic methods.}
	\label{figure:mix}
\end{figure*}

Results are shown in Figure~\ref{figure:mix} (d)-(j). The first row, with plots (d)-(f), shows comparison of Batch methods, where we see that VAN converges at a comparable rate to Newton's method.
The second row, with plots (h)-(j), shows the performance of the stochastic learning.
Since sVAN uses the full Hessian, it converges faster than sVAN-D and AdaGrad which use a diagonal approximation.
sVAN-D shows overall similar performance to AdaGrad.
The main advantage of sVAN over AdaGrad is that sVAN maintains a distribution $q(\btheta)$ which can be used to evaluate the uncertainty of the learned solution, as shown in the next subsection on active learning.

In conclusion, VAN and sVAN give comparable results to Newton's method while sVAN-D gives comparable results to AdaGrad.

\subsection{Active Learning for using VAN}
An important difference between active and stochastic learning is that an active learning agent can \emph{query} for its training data examples in each iteration.
In active learning for classification, examples in a pool of input data $\{\vx_i\}_{i=1}^{N}$ are ranked using an \emph{acquisition score} which measures how informative an example is for learning.
We pick the top $M$ data examples as the mini-batch.
Active learning is expected to be more data efficient than stochastic learning since the learning agent focuses on the most informative data samples.

In our experiments, we use the entropy score \citep{Schein2007} as the acquisition score to select data example for binary logistic regression:
\begin{align}
   \mathcal{H}(\vx) = - \sum_{c\in\{-1,1\}} \widehat{p}(y=c|\vx) \log \widehat{p}(y=c|\vx),
\end{align}
where $\widehat{p}(y=c|\vx)$ is the estimated probability that the label for input $\vx$ takes a value $y=c$. Within our VAN framework, we estimate these probabilities using distributions $q(\vtheta) = \gauss(\vtheta|\vmu_t,\vSigma_t)$ at iteration $t$. We use the following approximation that computes the probability using samples from $q$: 
\begin{align}
\widehat{p}_t(y=c|\vx) 
&\approx \int p(y=c|\vx, \btheta) q(\vtheta|\vmu_t,\vSigma_t) d\vtheta \\
&\approx \frac{1}{S} \sum_{s=1}^S p(y=c|\vx, \btheta_t^{(s)}),
\end{align}
where $S$ is the number of MC samples, $\vtheta_t^{(s)}$ are sample from $q$, and $p(y|\vx,\vtheta)$ is the logistic likelihood.

%Active learning requires specification of acquisition score.
%For classification, many acquisition functions are computed based on an uncertainty of the classifier which is computed by
%\begin{align}
%\widehat{p}(y|\bx) 
%&= \mathbb{E}_{q(\btheta)} \widehat{p}(y|\bx, \btheta)  \approx \frac{1}{U} \sum_{u=1}^U \widehat{p}(y|\bx, \btheta_u),
%\end{align}
%where $\{ \btheta_u \}_{u=1}^U \sim q(\btheta)$ and we use $U = 10$ in experiments.
%Then, the model uncertainty $\widehat{p}(y|\bx)$ is used to computed an acquisition score such as the \emph{entropy score}:
%\begin{align}
%\mathrm{acc}_{\mathrm{Ent}}(\bx) = -\sum_{c \in \mathcal{C}} \widehat{p}(y=c|\bx) \log \widehat{p}(y=c|\bx),
%\end{align}
%where $\mathcal{C}$ is the set of class labels.

\begin{figure*}[t]
	\centering
	\begin{minipage}{0.49\linewidth}
	\includegraphics[width=1\linewidth]{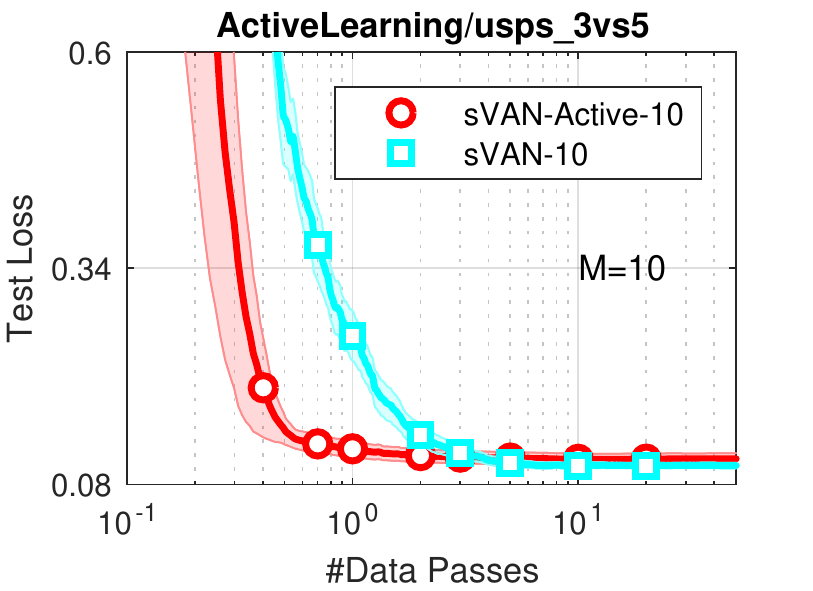}
	\vspace{-0.3cm}
	\caption{Active learning on logistic regression where we see that sVAN with active learning gives good results with fewer number of passes through the data. }
	\label{figure:active}
	\end{minipage}
\hfill
	\begin{minipage}{0.49\linewidth}
	\includegraphics[width=1\linewidth]{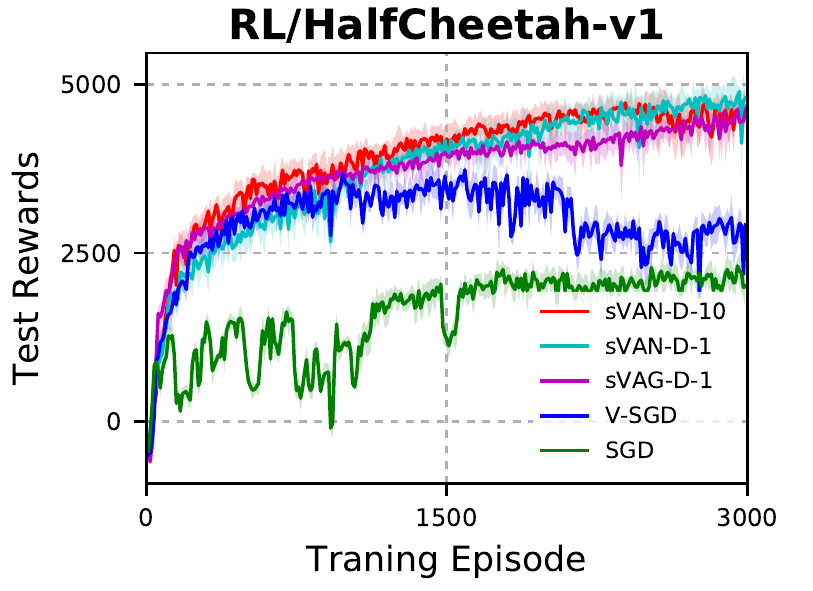}
	\vspace{-0.3cm}
	\caption{Parameter-based exploration for DDPG where we see than VAN based methods perform better than exploration using V-SGD or plain SGD with no exploration.}
	\label{figure:rl}
	\end{minipage}
\end{figure*}

Figure~\ref{figure:active} compares the performance on on the USPS dataset with active learning by VAN for mini-batch of 10 examples.
The result clearly shows that VAN with active learning is much more data efficient and stable than VAN with stochastic learning.
%Both method converge to good solutions while VAN with active learning seems to achieve a slightly worse solution.
%This is mainly because a simple acquisition score that we used does not prevent the same samples to be repeatedly queried and thus the active learning agent may overfit to the training data.
%In this experiment, we prevent this phenomena to some extent by removing the input data samples chosen in the last $20$ iterations from the pool of samples.
%However, we believe that a more sophisticated acquisition function is required for a robust active learning agent.

\subsection{Unsupervised Learning with Variational Auto-Encoder}
We apply VAN to optimize the parameters of variational auto-encoder (VAE) \citep{kingma2013auto}. 
Given observations $\{\by_i\}_{i=1}^N \sim p(\by)$, VAE models the data points using a neural-network decoder $p(\by_i|\bz_i,\vtheta_d)$ with input $\bz_i$ and parameters $\btheta_d$. The input $\bz_i$ are probabilistic and follow a prior $p(\vz_i)$. The encoder is also parameterized with a neural-network but follows a different distribution $q(\bz_i|\by_i,\vtheta_e)$ with parameters $\vtheta_e$. The goal is to learn $\vtheta := \{\vtheta_d,\vtheta_e\}$ by minimizing the following,
\begin{align}
   f(\vtheta) &= - \sum_{i=1}^N \myexpect_{q(\bz_i|\by_i,\btheta_e)} \sqr{ \log p(\by_i|\bz_i,\vtheta_d) } \notag \\
              &\quad\quad\quad\phantom{=} ~  \left. +\dkls{}{q(\bz_i|\by_i,\btheta_e)}{p(\bz_i)} \right] 
\end{align}
where $\bz_i$ is the latent vector, and $\vtheta=\{\vtheta_1, \vtheta_2 \}$ are parameters to be learned. Similar to previous work, we assume $p(\vz_i)$ to be a standard Gaussian, and use a Gaussian encoder and a Bernoulli decoder, both parameterized by neutral networks.

%Each distribution is parameterized by one hidden layer with $U = 100$ hidden units and $tanh$ activations. The encoder has two output layers that parametrizes the mean and the log-variance of the latent variables $\vz_n$. The decoder outputs the natural parameter of the Bernoulli distribution the models the observation $\vX$.
We train the model for the binary MNIST dataset ($N=70,000,D=256,N_{train}=60,000)$ with mini-batches of size $M = 100$ and set the dimensionality of the latent variable to $K = 3$. 
and measure the learning performance by the imputed log-loss of the test set using a procedure similar to \cite{rezende2015variational}. 

We compare our methods to adaptive-gradient methods, namely AdaGrad \citep{duchi2011adaptive} and RMSprop \citep{hintonTieleman}. For all the methods, we tune the step-size using a validation set and report the test log-loss after convergence.
Figure \ref{figure:mix} (c) shows the results of 5 runs of all methods. We see that all methods perform similar to each other. RMSprop has high variance among all methods. sVAN-D-1 is slightly worse than sVAG-D-1, which is expected since for nonconvex problems Gauss-Newton is a better approximation than using the Hessian.
%We record the test log-loss without imputation, which can be found in the appendix \ref{appendix:results}. In this case, all methods except sVAN-D-1 also perform similarly. 

\subsection{Parameter Based Exploration in Deep Reinforcement Learning}
Exploration is extremely useful in reinforcement learning (RL), especially in environment where obtaining informative feedback is rare. Unlike most learning tasks where a training dataset is readily available, RL agents needs to explore the state-action space to collect data. An explorative agent that always chooses random actions might never obtain good data for learning the optimal policy. On the other hand, an exploitative agent that always chooses the current optimal action(s) may never try suboptimal actions that can lead to learning a better policy. Thus, striking a balance between exploration and exploitation is a must for an effective learning. However, finding such a balance is an open problem in RL.

In this section, we apply VAN to enable efficient exploration by using the parameter-based exploration~\citep{DBLP:journals/paladyn/RuckstiessSSWSS10}.
In standard RL setup, we wish to learn a parameter $\btheta$ of a parametric policy $\vpi$ for action $\ba = \vpi(\bs; \btheta)$ given state $\bs$. We seek an optimal $\btheta$ such that a state-action sequence $(\bs_1, \ba_1, \bs_2 \dots)$ maximizes the expected returns where $\bs_t$ and $\ba_t$ are state and action at time $t$, respectively.
To facilitate exploration, a common approach is to perturb the action by a random noise $\vepsilon$, e.g., we can simply add it to the action $\ba =\vpi(\bs;\btheta) + \vepsilon$.
In contrast, in parameter-based exploration, exploration is facilitated in the parameter space as the name suggest, i.e., we sample the policy parameter $\vtheta$ from a distribution $\mathcal{N}(\btheta|\bmu, \bSigma)$.
Our goal therefore is to learn the distribution parameters $\bmu$ and $\bSigma$.

The parameter-based exploration is better than the action-space exploration because in the former a small perturbation of a parameter result in significant explorative behaviour in the action space. However, to get a similar behaviour through an action-space exploration, the extent of the noise needs to be very large which leads to instability \citep{DBLP:journals/paladyn/RuckstiessSSWSS10}.

The existing methods for parameter-based exploration independently learn $\bmu$ and $\bSigma$ (similar to V-SGD) \citep{DBLP:journals/paladyn/RuckstiessSSWSS10, plappert2017parameter,fortunato2017noisy}. However, this method can be increasingly unstable as learning progresses (as discussed in Section \ref{sec:vo}).
Since VAN exploits the geometry of the Gaussian distribution to jointly learn $\bmu$ and $\bSigma$, we expect it to perform better than the existing methods that use V-SGD.

We consider the \emph{deep deterministic policy gradient} (DDPG) method~\citep{DBLP:conf/icml/SilverLHDWR14, DBLP:journals/corr/LillicrapHPHETS15}, where a local optimal $\btheta$ is obtained by minimizing
\begin{align}
f(\btheta) = -\mathbb{E}_{p(\bs)}\left[ \widehat{Q}(\bs, \vpi(\bs,\btheta))\right],
\end{align}
where $\widehat{Q}(\bs, \ba)$ is an estimated of the expected return and $p(\bs)$ a state distribution.
Both $\vpi$ and $\widehat{Q}$ are neural networks.
We compare the performance of sVAN and sVAG against two baseline methods denoted by SGD and V-SGD.
SGD refers to DPG where there are no exploration at all, and V-SGD refers to an extension of DPG where the mean and covariance of the parameter distribution are learned using the update \eqref{eq:vsgd_mean} and \eqref{eq:vsgd_cov}.
Due to the large number of neural networks parameters, we use diagonal approximation to the Hessian for all methods.
More details of experiments are given in Appendix~\ref{appendix:rl}.

The result in Figure~\ref{figure:rl} shows the performance on Half-Cheetah from OpenAI Gym~\citep{1606.01540} for 5 runs, where we see that VAN based methods significantly outperform existing methods.
sVAN-D-1 and sVAG-D-1 both perform equally well. This suggests that both Hessian approximations obtained by using the reparameterization trick shown in \eqref{eq:reparam} and the Gauss-Newton approximation shown in \eqref{eq:Vag_prec}, respectively, are equally accurate for this problem.
We can also see that sVAN-D-10 has better data efficiency than sVAN-D-1 especially in the early stages of learning.
V-SGD is able to find a good policy during learning but has unstable performance that degenerates over time, as mentioned previously.
On the other hand, SGD performs very poorly and learns a suboptimal solution.
This strongly suggests that good exploration strategy is crucial to learn good policies of Half-Cheetah.

We also tried these comparisons on the `pendulum' problem in OpenAI Gym where we did not observe significant advantages from explorations. We believe that this is because this problem does not benefit from using exploration and pure-exploitative methods are good enough for these problems. More extensive experiments are required to validate the results presented in this section.

\section{Discussion and Conlcusions}
\label{sec:discussion}
%In Section \ref{sec:VANdiag}, we designed scalable versions of VAN using mean-field approximations. One could go structured approximation $q$, e.g., a chain or tree, which could enable a better approximation yet keeping the cost low. This is a future direction that could be a new way to design computationally-efficient approximations for second-order methods.

%%%%%%%%%%%%%%%%%%%%%%%
%% Content: We found %%
%%%%%%%%%%%%%%%%%%%%%%%
We proposed a general-purpose explorative-learning method called VAN. VAN is derived within the variational-optimization problem by using a mirror-descent algorithm. We showed that VAN is a second-order method and is related to existing methods in continuous optimization, variational inference, and evolution strategies. We proposed computationally-efficient versions of VAN for large-scale learning.
Our experimental results showed that VAN works reasonably well on a wide-variety of learning problems. 

For problems with high-dimensional parameters $\vtheta$, computing and inverting the full Hessian matrix is computationally infeasible. 
One line of possible future work is to develop versions of VAN that can deal with issue without making a diagonal approximation.

It is straightforward to extend VAN to non-Gaussian distributions. Our initial work, not discussed in this paper, suggests that student-t distribution and Cauchy distribution could be useful candidate. However, it is always possible to use other types of distributions, for example, to minimize discrete optimization problem within a stochastic relaxation framework \citep{geman1984stochastic}.
%Such distributions will be useful, for example, to design methods that can avoid pathological flat regions and local minima in problems such as deep learning.

Another venue is to evaluate the impact of exploration on fields such as active learning and reinforcement learning. In this paper, we have provided some initial results. An extensive application of the methods developed in this paper to real-world problems is required to further understand their advantages and disadvantages compared to existing methods.
%
%Another possible area where VAN can make a difference is variational inference. We have only commented on the connection between the two fields. VAN views $q$ as an explorative mechanism which is a new way to interpret variational inference methods. This new insight can be used to design new methods for variational inference. The experimental results in this paper show that this is possible.

The main strength of VAN lies in exploration, using which it can potentially accelerate and robustify optimization. Compared to Bayesian methods, VAN offers a computationally cheap alternative to perform explorative learning. Using such cheap explorations, VAN has the potential to solve difficult learning problems such as deep reinforcement learning, active learning, and life-long learning. 

%VAN is a form of variational optimization, which has several benefits.
%Variational optimization provides an alternative way to optimize the original objective function by exploration.
%Compared to existing variational optimization methods, our method gives simple, efficient natural gradient updates for both $\vmu$ and $\vSigma$.

%In this paper, we focused only on the case of Gaussian $q(\vtheta)$. It can therefore not be used for discrete or constrained optimization problems. However, the method could be generalized to other exponential family distributions.

\bibliographystyle{apalike}
\bibliography{paper}

\begin{thebibliography}{}

\bibitem[Amari, 1998]{amari1998natural}
Amari, S.-I. (1998).
\newblock Natural gradient works efficiently in learning.
\newblock {\em Neural computation}, 10(2):251--276.

\bibitem[Bertsekas, 1999]{bertsekas1999nonlinear}
Bertsekas, D.~P. (1999).
\newblock {\em Nonlinear programming}.
\newblock Athena Scientific.

\bibitem[Bhatnagar, 2007]{bhatnagar2007adaptive}
Bhatnagar, S. (2007).
\newblock Adaptive {N}ewton-based multivariate smoothed functional algorithms
  for simulation optimization.
\newblock {\em ACM Transactions on Modeling and Computer Simulation (TOMACS)},
  18(1):2.

\bibitem[Bishop, 2006]{bishop2006pattern}
Bishop, C.~M. (2006).
\newblock Pattern recognition.
\newblock {\em Machine Learning}, 128:1--58.

\bibitem[Brochu et~al., 2010]{brochu2010tutorial}
Brochu, E., Cora, V.~M., and De~Freitas, N. (2010).
\newblock A tutorial on {B}ayesian optimization of expensive cost functions,
  with application to active user modeling and hierarchical reinforcement
  learning.
\newblock {\em arXiv preprint arXiv:1012.2599}.

\bibitem[Brockman et~al., 2016]{1606.01540}
Brockman, G., Cheung, V., Pettersson, L., Schneider, J., Schulman, J., Tang,
  J., and Zaremba, W. (2016).
\newblock Openai gym.

\bibitem[Byrd et~al., 2016]{byrd2016stochastic}
Byrd, R.~H., Hansen, S.~L., Nocedal, J., and Singer, Y. (2016).
\newblock A stochastic quasi-{N}ewton method for large-scale optimization.
\newblock {\em SIAM Journal on Optimization}, 26(2):1008--1031.

\bibitem[Chang and Lin, 2011]{CC01a}
Chang, C.-C. and Lin, C.-J. (2011).
\newblock {LIBSVM}: A library for support vector machines.
\newblock {\em ACM Transactions on Intelligent Systems and Technology},
  2:27:1--27:27.
\newblock Software available at \url{http://www.csie.ntu.edu.tw/~cjlin/libsvm}.

\bibitem[Crammer et~al., 2009]{crammer2009adaptive}
Crammer, K., Kulesza, A., and Dredze, M. (2009).
\newblock Adaptive regularization of weight vectors.
\newblock In {\em Advances in neural information processing systems}, pages
  414--422.

\bibitem[Duchi et~al., 2011]{duchi2011adaptive}
Duchi, J., Hazan, E., and Singer, Y. (2011).
\newblock Adaptive subgradient methods for online learning and stochastic
  optimization.
\newblock {\em The Journal of Machine Learning Research}, 12:2121--2159.

\bibitem[Fishel and Loeb, 2012]{fishel2012bayesian}
Fishel, J.~A. and Loeb, G.~E. (2012).
\newblock Bayesian exploration for intelligent identification of textures.
\newblock {\em Frontiers in neurorobotics}, 6.

\bibitem[Fortunato et~al., 2017]{fortunato2017noisy}
Fortunato, M., Azar, M.~G., Piot, B., Menick, J., Osband, I., Graves, A., Mnih,
  V., Munos, R., Hassabis, D., Pietquin, O., et~al. (2017).
\newblock Noisy networks for exploration.
\newblock {\em arXiv preprint arXiv:1706.10295}.

\bibitem[{Gal} et~al., 2017]{2017arXiv170302910G}
{Gal}, Y., {Islam}, R., and {Ghahramani}, Z. (2017).
\newblock {Deep Bayesian Active Learning with Image Data}.
\newblock {\em ArXiv e-prints}.

\bibitem[Geman and Geman, 1984]{geman1984stochastic}
Geman, S. and Geman, D. (1984).
\newblock Stochastic relaxation, {G}ibbs distributions, and the {B}ayesian
  restoration of images.
\newblock {\em IEEE Transactions on pattern analysis and machine intelligence},
  1(6):721--741.

\bibitem[Ghadimi et~al., 2014]{ghadimi2014mini}
Ghadimi, S., Lan, G., and Zhang, H. (2014).
\newblock Mini-batch stochastic approximation methods for nonconvex stochastic
  composite optimization.
\newblock {\em Mathematical Programming}, pages 1--39.

\bibitem[G{\"u}rb{\"u}zbalaban et~al., 2015]{gurbuzbalaban2015globally}
G{\"u}rb{\"u}zbalaban, M., Ozdaglar, A., and Parrilo, P. (2015).
\newblock A globally convergent incremental {N}ewton method.
\newblock {\em Mathematical Programming}, 151(1):283--313.

\bibitem[Hansen and Ostermeier, 2001]{hansen2001completely}
Hansen, N. and Ostermeier, A. (2001).
\newblock Completely derandomized self-adaptation in evolution strategies.
\newblock {\em Evolutionary computation}, 9(2):159--195.

\bibitem[{Houlsby} et~al., 2011]{2011arXiv1112.5745H}
{Houlsby}, N., {Husz{\'a}r}, F., {Ghahramani}, Z., and {Lengyel}, M. (2011).
\newblock {Bayesian Active Learning for Classification and Preference
  Learning}.
\newblock {\em ArXiv e-prints}.

\bibitem[Huszar, 2017]{ferencBlogES}
Huszar, F. (2017).
\newblock {Evolution Strategies, Variational Optimisation and Natural ES}.
\newblock {\tiny
  \url{http://www.inference.vc/evolution-strategies-variational-optimisation-and-natural-es-2/}}.

\bibitem[Khan and Lin, 2017]{khan2017conjugate}
Khan, M.~E. and Lin, W. (2017).
\newblock Conjugate-computation variational inference: Converting variational
  inference in non-conjugate models to inferences in conjugate models.
\newblock {\em arXiv preprint arXiv:1703.04265}.

\bibitem[Kingma and Ba, 2014]{kingma2014adam}
Kingma, D. and Ba, J. (2014).
\newblock {Adam: A method for stochastic optimization}.
\newblock {\em arXiv preprint arXiv:1412.6980}.

\bibitem[Kingma and Welling, 2013]{kingma2013auto}
Kingma, D.~P. and Welling, M. (2013).
\newblock Auto-encoding variational {Bayes}.
\newblock {\em arXiv preprint arXiv:1312.6114}.

\bibitem[Lillicrap et~al., 2015]{DBLP:journals/corr/LillicrapHPHETS15}
Lillicrap, T.~P., Hunt, J.~J., Pritzel, A., Heess, N., Erez, T., Tassa, Y.,
  Silver, D., and Wierstra, D. (2015).
\newblock Continuous control with deep reinforcement learning.
\newblock {\em CoRR}, abs/1509.02971.

\bibitem[Marlin et~al., 2011]{marlin2011piecewise}
Marlin, B., Khan, M., and Murphy, K. (2011).
\newblock Piecewise bounds for estimating {B}ernoulli-logistic latent
  {G}aussian models.
\newblock In {\em International Conference on Machine Learning}.

\bibitem[Mokhtari et~al., 2016]{mokhtari2016adaptive}
Mokhtari, A., Daneshmand, H., Lucchi, A., Hofmann, T., and Ribeiro, A. (2016).
\newblock Adaptive {N}ewton method for empirical risk minimization to
  statistical accuracy.
\newblock In {\em Advances in Neural Information Processing Systems}, pages
  4062--4070.

\bibitem[Opper and Archambeau, 2009]{Opper:09}
Opper, M. and Archambeau, C. (2009).
\newblock {The Variational {G}aussian Approximation Revisited}.
\newblock {\em Neural Computation}, 21(3):786--792.

\bibitem[Perfors et~al., 2011]{perfors2011tutorial}
Perfors, A., Tenenbaum, J.~B., Griffiths, T.~L., and Xu, F. (2011).
\newblock A tutorial introduction to {B}ayesian models of cognitive
  development.
\newblock {\em Cognition}, 120(3):302--321.

\bibitem[Plappert et~al., 2017]{plappert2017parameter}
Plappert, M., Houthooft, R., Dhariwal, P., Sidor, S., Chen, R.~Y., Chen, X.,
  Asfour, T., Abbeel, P., and Andrychowicz, M. (2017).
\newblock Parameter space noise for exploration.
\newblock {\em arXiv preprint arXiv:1706.01905}.

\bibitem[Ranganath et~al., 2014]{ranganath2013black}
Ranganath, R., Gerrish, S., and Blei, D.~M. (2014).
\newblock Black box variational inference.
\newblock In {\em International conference on Artificial Intelligence and
  Statistics}, pages 814--822.

\bibitem[Raskutti and Mukherjee, 2015]{raskutti2015information}
Raskutti, G. and Mukherjee, S. (2015).
\newblock The information geometry of mirror descent.
\newblock {\em IEEE Transactions on Information Theory}, 61(3):1451--1457.

\bibitem[Rezende and Mohamed, 2015]{rezende2015variational}
Rezende, D.~J. and Mohamed, S. (2015).
\newblock Variational inference with normalizing flows.
\newblock {\em arXiv preprint arXiv:1505.05770}.

\bibitem[R{\"{u}}ckstie{\ss} et~al.,
  2010]{DBLP:journals/paladyn/RuckstiessSSWSS10}
R{\"{u}}ckstie{\ss}, T., Sehnke, F., Schaul, T., Wierstra, D., Sun, Y., and
  Schmidhuber, J. (2010).
\newblock Exploring parameter space in reinforcement learning.
\newblock {\em Paladyn}, 1(1):14--24.

\bibitem[{Salimans} et~al., 2017]{2017arXiv170303864S}
{Salimans}, T., {Ho}, J., {Chen}, X., {Sidor}, S., and {Sutskever}, I. (2017).
\newblock {Evolution Strategies as a Scalable Alternative to Reinforcement
  Learning}.
\newblock {\em ArXiv e-prints}.

\bibitem[Schein and Ungar, 2007]{Schein2007}
Schein, A.~I. and Ungar, L.~H. (2007).
\newblock Active learning for logistic regression: an evaluation.
\newblock {\em Machine Learning}, 68(3):235--265.

\bibitem[Silver et~al., 2014]{DBLP:conf/icml/SilverLHDWR14}
Silver, D., Lever, G., Heess, N., Degris, T., Wierstra, D., and Riedmiller,
  M.~A. (2014).
\newblock Deterministic policy gradient algorithms.
\newblock In {\em Proceedings of the 31th International Conference on Machine
  Learning, {ICML} 2014, Beijing, China, 21-26 June 2014}, pages 387--395.

\bibitem[Spall, 2000]{spall2000adaptive}
Spall, J.~C. (2000).
\newblock Adaptive stochastic approximation by the simultaneous perturbation
  method.
\newblock {\em IEEE transactions on automatic control}, 45(10):1839--1853.

\bibitem[{Staines} and {Barber}, 2012]{2012arXiv1212.4507S}
{Staines}, J. and {Barber}, D. (2012).
\newblock {Variational Optimization}.
\newblock {\em ArXiv e-prints}.

\bibitem[Strens, 2000]{Strens00abayesian}
Strens, M. (2000).
\newblock A {B}ayesian framework for reinforcement learning.
\newblock In {\em In Proceedings of the Seventeenth International Conference on
  Machine Learning}, pages 943--950. ICML.

\bibitem[Tieleman and Hinton, 2012]{hintonTieleman}
Tieleman, T. and Hinton, G. (2012).
\newblock {Lecture 6.5-{R}MSprop: Divide the gradient by a running average of
  its recent magnitude.}
\newblock {\em COURSERA: Neural Networks for Machine Learning 4}.

\bibitem[Wainwright and Jordan, 2008]{WainwrightJordan08}
Wainwright, M.~J. and Jordan, M.~I. (2008).
\newblock Graphical models, exponential families, and variational inference.
\newblock {\em Foundations and Trends in Machine Learning}, 1--2:1--305.

\bibitem[Wierstra et~al., 2008]{wierstra2008natural}
Wierstra, D., Schaul, T., Peters, J., and Schmidhuber, J. (2008).
\newblock Natural evolution strategies.
\newblock In {\em Evolutionary Computation, 2008. CEC 2008.(IEEE World Congress
  on Computational Intelligence). IEEE Congress on}, pages 3381--3387. IEEE.

\bibitem[Williams, 1992]{williams1992simple}
Williams, R.~J. (1992).
\newblock Simple statistical gradient-following algorithms for connectionist
  reinforcement learning.
\newblock {\em Machine learning}, 8(3-4):229--256.

\bibitem[Wyatt, 1998]{wyatt1998exploration}
Wyatt, J. (1998).
\newblock {\em Exploration and inference in learning from reinforcement}.
\newblock PhD thesis, University of Edinburgh. College of Science and
  Engineering. School of Informatics.

\end{thebibliography}

\newpage
\onecolumn
\begin{appendix}
    \section{Derivation of VAN}
\label{appendix:derivation_van}

Denote the mean parameters of $q_t(\vtheta)$ by $\vm_t$ which is equal to the expected value of the sufficient statistics $\vphi(\vtheta)$, i.e., $\vm_t := \myexpect_{q_t} [\vphi(\vtheta)] $. The mirror descent update at iteration $t$ is given by the solution to
\begin{align}
\vm_{t+1} 
&= \argmin_{\boldsymbol{m}} \left\langle \vm, \widehat{\nabla}_{m} \mathcal{L}_t \right\rangle + \frac{1}{\beta_t}\dkls{}{q}{q_t} \\
&= \argmin_{\boldsymbol{m}} \myexpect_q \left[ \left\langle \vphi(\vtheta), \widehat{\nabla}_{m} \mathcal{L}_t \right\rangle + \log\left((q/q_t)^{1/\beta_t}\right) \right] \\
&= \argmin_{\boldsymbol{m}} \myexpect_q \left[ \log \frac{ \exp \left\langle \vphi(\vtheta), \widehat{\nabla}_{m} \mathcal{L}_t \right\rangle q^{1/\beta_t} }{ q_t^{1/\beta_t}} \right] \\
&= \argmin_{\boldsymbol{m}} \myexpect_q \left[ \log \left( \frac{q^{1/\beta_t}}{q_t^{1/\beta_t} \exp \left\langle \vphi(\vtheta), -\widehat{\nabla}_{m} \mathcal{L}_t \right\rangle } \right) \right] \\
&= \argmin_{\boldsymbol{m}} \frac{1}{\beta_t} \,\, \myexpect_q \left[ \log \left( \frac{q}{q_t \exp \left\langle \vphi(\vtheta), -\beta_t \widehat{\nabla}_{m} \mathcal{L}_t \right\rangle } \right) \right] \\
&= \argmin_{\boldsymbol{m}}  \frac{1}{\beta_t} \,\,\mathbb{D}_{KL}\sqr{ q \| q_t \exp \left\langle \vphi(\vtheta), -\beta_t \widehat{\nabla}_{m} \mathcal{L}_t \right\rangle /\mathcal{Z}_t} .
\end{align}
where $\mathcal{Z}$ is the normalizing constant of the distribution in the denominator which is a function of the gradient and step size.

Minimizing this KL divergence gives the update
\begin{equation}
q_{t+1}(\vtheta) \propto q_t(\vtheta)\exp \left\langle \vphi(\vtheta), -\beta_t \widehat{\nabla}_{m} \mathcal{L}_t \right\rangle.
\end{equation}
By rewriting this, we see that we get an update in the natural parameters $\vlambda_t$ of $q_t(\vtheta)$, i.e.
\begin{equation}
\vlambda_{t+1} = \vlambda_t - \beta_t \widehat{\nabla}_{m} \mathcal{L}_t.
\end{equation}
Recalling that the mean parameters of a Gaussian $q(\vtheta) = \gauss(\vtheta|\vmu,\vSigma)$ are $\vm^{(1)} = \vmu$ and $\vM^{(2)} = \vSigma + \vmu\vmu ^T$ and using the chain rule, we can express the gradient $\widehat{\nabla}_{m} \mathcal{L}_t$ in terms of $\vmu$ and $\vSigma$,
\begin{align}
    \widehat{\nabla}_{m^{(1)}} \mathcal{L} &= \widehat{\nabla}_{\mu} \mathcal{L} - 2\left[\widehat{\nabla}_{\Sigma} \mathcal{L} \right]\vmu \\
    \widehat{\nabla}_{M^{(2)}} \mathcal{L} &= \widehat{\nabla}_{\Sigma} \mathcal{L}.
\end{align}
Finally, recalling that the natural parameters of a Gaussian $q(\vtheta) = \gauss(\vtheta|\vmu,\vSigma)$ are $\vlambda^{(1)} = \vSigma^{-1}\vmu$ and $\vlambda^{(2)} = - \frac{1}{2}\vSigma^{-1}$, we can rewrite the VAN updates in terms of $\vmu$ and $\vSigma$,
\begin{align}
    \vSigma_{t+1}^{-1} &= \vSigma_t^{-1} + 2\beta_t \left[ \widehat{\nabla}_{\Sigma} \mathcal{L}_t \right] \\
    \vmu_{t+1} &= \vSigma_{t+1} \left[ \vSigma_t^{-1}\vmu_t - \beta_t \left( \widehat{\nabla}_{\mu} \mathcal{L}_t - 2\left[\widehat{\nabla}_{\Sigma} \mathcal{L}_t \right] \vmu_t \right) \right] \\
    &= \vSigma_{t+1} \left( \vSigma_t^{-1} + 2\beta_t \left[ \widehat{\nabla}_{\Sigma} \mathcal{L}_t \right] \right)\vmu_t - \beta_t\vSigma_{t+1} \left[ \widehat{\nabla}_{\mu} \mathcal{L}_t \right] \\
    &= \vmu_t - \beta_t\vSigma_{t+1} \left[ \widehat{\nabla}_{\mu} \mathcal{L}_t \right].
\end{align}

\section{Details on the RL experiment}
\label{appendix:rl}

In this section, we give details of the parameter-based exploration task in reinforcement learning (RL). An important open question in reinforcement learning is how to efficiently explore the state and action space. An agent always acting greedily according to the policy results in a pure exploitation. Exploration is necessary to visit inferior state and actions once in while to see if they might really be better. 
Traditionally, exploration is performed in the action space by, e.g., injecting noise to the policy output.
However, injecting noise to the action space may not be sufficient in problems where the reward is sparse, i.e., the agent rarely observes the reward of their actions.
In such problems, the agent requires a rich explorative behaviour in which noises in the action space cannot provide.
An alternative approach is to perform exploration in the \emph{parameter space}~\citep{DBLP:journals/paladyn/RuckstiessSSWSS10}.
In this section, we demonstrate that variational distribution $q(\vtheta)$ obtained using VAN can be straightforwardly used for such exploration in parameter space, $\vtheta$.

\subsection{Background}

First, we give a brief background on reinforcement learning (RL).
RL aims to solve the sequential decision making problem where at each discrete time step $t$ an agent observes a state $\vs_t$ and selects an action $\va_t$ using a policy $\pi$, i.e., $\va_t \sim \pi(\va|\vs_t)$. 
The agent then receives an immediate reward $r_t = r(\vs_t,\va_t)$ and observes a next state $\vs_t \sim p(\vs'|\vs_t, \va_t)$.
The goal in RL is to learn the optimal policy $\pi^*$ which maximizes the expected returns $\mathbb{E}\left[ \sum_{t}^\infty \gamma^{t-1} r_t \right]$ where $\gamma$ is the discounted factor and the expectation is taken over a sequence of densities $\pi(\va|\vs_t)$ and $p(\vs'|\vs_t, \va_t)$.

A central component of RL algorithms is the state-action value function or the Q-function $Q^{\pi}(\vs,\va)$ gives the expected return after executing an action $\va$ in a state $\vs$ and following the policy $\pi$ afterwards. Formally, it is defined as follows:
\begin{align}
Q^{\pi}(\vs,\va) = \mathbb{E}\left[ \sum_{t=1}^\infty \gamma^{t-1} r_t | \vs_1 = \vs, \va_1 = \va\right].
\end{align}
The Q-function also satisfies a recursive relation also known as the Bellman equation:
\begin{align}
Q^{\pi}(\vs,\va) = r(\vs,\va) + \gamma \mathbb{E}_{p(\vs'|\vs,\va), \pi(\va'|\vs')} \left[ Q^{\pi}(\vs', \va') \right].
\end{align}
Using the Q-function, the goal of reinforcement learning can be simply stated as finding a policy which maximizes the expected Q-function, i.e.,
\begin{align}
\pi^* = \argmax_{\pi} \iint p(\vs) \pi(\va|\vs) Q^{\pi}(\vs,\va) \ds \da. \label{eq:rl_q_objective}
\end{align}
In practice, the policy is represented by a parameterized function such as neural networks with policy parameter $\btheta$ and the goal is to instead find the optimal parameters $\btheta^\star$.

\subsubsection{Deterministic Policy Gradients}

Our parameter-based exploration via VAN can be applied to any reinforcement learning algorithms which rely on gradient ascent to optimize the policy parameter $\btheta$.
For demonstration, we focus on a simple yet efficient algorithm called the deterministic policy gradients algorithm (DPG)~\citep{DBLP:conf/icml/SilverLHDWR14}.
Simply speaking, DPG aims to find a deterministic policy that maximizes the action-value function by gradient ascent.
Since in practice the action-value function is unknown, DPG learns a function approximator $Q_{\bv}(\vs,\va)$ with a parameter $\bv$ such that $Q_{\bv}(\vs,\va) \approx Q^{\pi_{\btheta}}(\vs,\va)$.
Then, DPG finds $\btheta^*$ which locally minimize an objective $f(\btheta) = -\mathbb{E}_{p(\vs)} \left[ Q_{\bv}(\vs, \pi_{\btheta}(\vs)) \right]$ by gradient ascent where the gradient is given by
\begin{align}
\nabla_{\btheta} f(\btheta)
&= -\nabla_{\btheta} \int p(\vs) Q_{\bv}(\vs,\pi_{\btheta}(\vs)) \ds \notag \\
&= -\int p(\vs) \nabla_{\btheta} Q_{\bv}(\vs,\pi_{\btheta}(\vs)) \ds \notag \\
&= -\int p(\vs) \nabla_{\va} Q_{\bv}(\vs,\va)|_{\va = \pi_{\btheta}(\vs)} \nabla_{\btheta} \pi_{\btheta}(\vs) \ds. \label{eq:dpg_update}
\end{align}

The parameter $\bv$ of $Q_{\bv}(\vs,\va)$ may be learned by any policy evaluation methods.
Here, we adopted the approach proposed by~\citep{DBLP:journals/corr/LillicrapHPHETS15} which minimizes the squared Bellman residual to the slowly moving target action-value function.
More precisely, $\bv$ is updated by 
\begin{align}
\min_{\bv} \mathbb{E}\left[ \left( Q_{\bv}(\vs,\va) - r(\vs,\va) - \gamma \tilde{Q}_{\tilde{\bv}}(\vs', \tilde{\pi}_{\tilde{\btheta}}(\vs'))\right)^2  \right],
\label{eq:q_residual}
\end{align}
where the expectation is taken over $p(\vs)$ and $p(\vs'|\vs,\va)$.
The $\tilde{Q}_{\tilde{\bv}}$ and $\tilde{\pi}_{\tilde{\btheta}}$ denote target networks which are  separate function approximators that slowly tracks $Q_{\bv}$ and $\pi_{\btheta}$, respectively. The target networks help at stabilizing the learning procedure~\citep{DBLP:journals/corr/LillicrapHPHETS15}.

Overall, DPG is an actor-critic algorithm that iteratively update the critic (action-value function) by taking gradient of Eq.\eqref{eq:q_residual} and update the actor (policy) by the gradient Eq.\eqref{eq:dpg_update}.
However, the crucial issue of DPG is that it uses a deterministic policy and does not perform exploration by itself. 
In practice, exploration is done for DPG by injecting a noise to the policy output, i.e., $\va = \pi_{\btheta}(\vs) + \boldsymbol{\epsilon}$ where $\boldsymbol{\epsilon}$ is a noise from some random process such as Gaussian noise.
However, as discussed about, action-space noise may be insufficient in some problems.
Next, we show that VAN can be straightforwardly applied to DPG to obtain \emph{parameter-based exploration DPG}.

\subsubsection{Parameter-based Exploration DDPG}
To perform parameter-based exploration, we can relax the policy-gradient objective $f(\btheta)$ by assuming that the parameter $\btheta$ is sampled from a distribution $q(\btheta) := \mathcal{N}(\btheta | \bmu, \bSigma)$, and solve the following optimization problem: 
\begin{align}
   \min_{\bmu,\bSigma} \,\, \myexpect_{\mathcal{N}(\btheta|\bmu,\bSigma)}[f(\btheta)]
\end{align}
This is exactly the VO problem of \eqref{eq:VO}.
The stochasticity of $\btheta$ through $q(\vtheta)$ allows the agent to explore the state and action space by varying its policy parameters.
This exploration strategy is advantageous since the agent can now exhibits much more richer explorative behaviours when compared with exploration by action noise injection.

   Algorithm \ref{algorithm:dpg_van} outlines \emph{parameter-based exploration DPG via VAN}.
   \begin{algorithm}[t]
      \begin{algorithmic}[1]
   	\WHILE {Not converged}
   		\STATE Observe state $\vs_t$, sample parameter $\btheta \sim \mathcal{N}(\btheta | \bmu, \bSigma)$, take action $\va_t = \pi_{\btheta}(\vs_t)$, observe reward $r_t$ and next state $\vs'_t$.
   		\STATE Add $(\vs_t, \va_t, r_t, \vs'_t)$ to a replay buffer $\mathcal{D}$.
   		\FOR {$i = 1,\dots,K$}
   			\STATE Drawn N minibatch samples $\{(\vs_i, \va_i, r_i, \vs'_i)\}_{i=1}^N$ from $\mathcal{D}$.
   			\STATE Update $Q_{\bv}(\vs,\va)$ by gradient descent:
   			\begin{align*}
   			\bv \leftarrow \bv + \alpha_{\bv} \nabla_{\bv} \frac{1}{N} \sum_{i=1}^N 
   			\left[ \left( Q_{\bv}(\vs_i,\va_i) - y_i\right)^2  \right],
   			\end{align*}
   			where $y_i = r(\vs_i,\va_i) + \gamma \tilde{Q}_{\tilde{\bv}}(\vs'_i, \pi_{\tilde{\bmu}}(\vs'_i))$.
   			\STATE Update parameter of $q(\btheta)$ by sVAN in Eq.\eqref{eq:Vand_prec_0}  with
   			\begin{align}
               \widehat{\nabla}_{\sigma} \mathcal{L}(\bmu_t, \vsigma_t^2) 
   			&= - \frac{1}{NM}\sum_{i=1}^N\sum_{j=1}^M \left[ \nabla_{\sigma}\pi_{\btheta_j}(\vs_i) \nabla_{\ba} Q_{\bv}(\vs_i,\ba)|_{\ba = \pi_{\btheta_j}(\vs_i)} \right], \\
   			\widehat{\nabla}_{\bmu} \mathcal{L}(\bmu_t, \vsigma_t^2) 
   			&= - \frac{1}{NM}\sum_{i=1}^N\sum_{j=1}^M \left[ \nabla_{\bmu}\pi_{\btheta_j}(\vs_i) \nabla_{\ba} Q_{\bv}(\vs_i,\ba)|_{\ba = \pi_{\btheta_j}(\vs_i)} \right],
   			\end{align}
   			where $\{\btheta_j \}_{j=1}^M \sim q(\btheta)$.

   			\STATE Update target network parameters $\tilde{\bv}$ and $\tilde{\bmu}$ by moving average with, e.g., step size $\tau = 0.001$:
   			\begin{align*}
   			\tilde{\bv} &\leftarrow (1 - \tau) \tilde{\bv} + \tau \bv, \\	
   			\tilde{\bmu} &\leftarrow (1 - \tau) \tilde{\bmu} + \tau \bmu.
   			\end{align*}
   		\ENDFOR
   	\ENDWHILE
   	\end{algorithmic} 
   \caption{Parameter-based exploration DPG via VAN}
   \label{algorithm:dpg_van}
   \end{algorithm}

\end{appendix}

\end{document}